\documentclass{article}
\PassOptionsToPackage{numbers, sort&compress}{natbib}

\usepackage[final]{neurips_2025}

\usepackage[utf8]{inputenc} 
\usepackage[T1]{fontenc}    
\usepackage{hyperref}       
\usepackage{url}            
\usepackage{booktabs}       
\usepackage{amsmath}
\usepackage{amssymb}
\usepackage{amsthm}
\usepackage{amsfonts}       
\usepackage{nicefrac}       
\usepackage{microtype}      
\usepackage{xcolor}         
\usepackage[nolist]{acronym}
\usepackage{tikz}
\usetikzlibrary{automata,positioning}
\usepackage{algorithm}
\usepackage{algpseudocode}
\usepackage{caption}    
\usepackage{subcaption}
\usepackage{multirow}
\usepackage{mathtools}
\usepackage{bbm}
\usepackage{siunitx}
\usepackage{etoolbox}
\usepackage{bm}
\usepackage{wrapfig}
\usepackage{afterpage}

\newcommand{\expectation}{\mathop{\mathbb{E}}}

\DeclareMathOperator*{\argmax}{arg\,max}

\newtheorem{lemma}{Lemma}
\newtheorem{proposition}{Proposition}
\newtheorem{assumption}{Assumption}

\newenvironment{proofsketch}{%
  \proof}{\endproof}

\title{Automaton Constrained Q-Learning}

\author{%
  Anastasios Manganaris, Vittorio Giammarino, and Ahmed H. Qureshi \\
  Department of Computer Science\\
  Purdue University\\
  \texttt{\{amangana,vgiammar,ahqureshi\}@purdue.edu} \\
}

\begin{document}

\maketitle

\begin{abstract}

Real-world robotic tasks often require agents to achieve sequences of goals while respecting time-varying safety constraints. However, standard \ac{RL} paradigms are fundamentally limited in these settings. A natural approach to these problems is to combine \ac{RL} with \ac{LTL}, a formal language for specifying complex, temporally extended tasks and safety constraints. Yet, existing \ac{RL} methods for \ac{LTL} objectives exhibit poor empirical performance in complex and continuous environments. As a result, no scalable methods support both temporally ordered goals and safety simultaneously, making them ill-suited for realistic robotics scenarios.
We propose \ac{ACQL}, an algorithm that addresses this gap by combining goal-conditioned value learning with automaton-guided reinforcement. \ac{ACQL} supports most \ac{LTL} task specifications and leverages their automaton representation to explicitly encode stage-wise goal progression and both stationary and non-stationary safety constraints.
We show that \ac{ACQL} outperforms existing methods across a range of continuous control tasks, including cases where prior methods fail to satisfy either goal-reaching or safety constraints. We further validate its real-world applicability by deploying \ac{ACQL} on a 6-DOF robotic arm performing a goal-reaching task in a cluttered, cabinet-like space with safety constraints. Our results demonstrate that \ac{ACQL} is a robust and scalable solution for learning robotic behaviors according to rich temporal specifications.
\end{abstract}


%

\section{Introduction}

Achieving desirable robot behavior in real-world applications often requires managing long and complex sequences of subgoals while adhering to strict safety constraints. For instance, autonomous mobile robots in warehouse settings must navigate to restock shelves, all while avoiding obstacles and remaining within a reachable distance from the recharging station for their current battery level. Similarly, a nursing robot must routinely complete its rounds in a given sequence, unless notified of an emergency situation.


Despite the success of \ac{GCRL} \citep{schaul2015uvfa, andrychowicz2017her} and Safe \ac{RL} \citep{altman1999constrained,achiam2017cpo}, these approaches fall short in addressing the combined challenges of sequential goals and dynamic safety constraints. \ac{GCRL} is effective for reaching individual goals but lacks mechanisms for reasoning over goal sequences or ensuring safety. Safe \ac{RL} enforces fixed safety constraints, but typically assumes they remain static throughout the task, limiting its applicability to temporally evolving tasks.


Formal languages such as \ac{LTL} \cite{pnueli1977ltl} provide an expressive framework for specifying complex tasks involving multiple interdependent goals and non-stationary safety constraints. However, reward functions defined directly from \ac{LTL} specifications are inherently non-Markovian, since temporal properties refer to entire trajectories rather than individual transitions. This violates the Markov assumption---a fundamental premise of standard \ac{RL}---which assumes that transitions and rewards depend only on the current state and action. As a result, RL algorithms built on the \ac{MDP} formalism \cite{haarnoja2018soft, schulman2017proximalpolicyoptimizationalgorithms} are ill-suited for direct application to \ac{LTL}-defined objectives.

To overcome this, some studies \citep{camacho2019ltl, hasanbeig2020cautious, hasanbeig2020deep} have proposed to translate the \ac{LTL} formula into an automaton that tracks temporal progress toward task completion and enables defining Markovian rewards corresponding to temporal specifications. While this approach allows for potentially handling arbitrary \ac{LTL} formulas, its generality is tied to the use of sparse binary rewards derived from the Boolean evaluation of the specification. In complex environments, however, such sparse signals are encountered too infrequently to effectively guide behavior, and designing denser rewards typically requires additional domain knowledge not readily available in many robotics tasks. Furthermore, these methods only implicitly deal with ensuring safety, as safety properties are fundamental components of many \ac{LTL} specifications \cite{alpern1987Recognizing}, and typically rely on ad-hoc mechanisms, such as halting rollouts \cite{toroicarte2018RewardMachines} and reward shaping \cite{araki2021logicaloptionsframework}, both of which tend to be brittle and ineffective in complex domains. Other approaches seek to improve scalability by employing hierarchical~\cite{araki2021logicaloptionsframework} or goal-conditioned~\cite{qiu2023instructing} policies but sacrifice generality, particularly with respect to safety constraints.





To bridge the gap between \ac{LTL}-capable algorithms that scale poorly and scalable \ac{RL} algorithms that lack support for most \ac{LTL} tasks, we propose \ac{ACQL}, which lifts Safe \ac{RL} and \ac{GCRL} to the class of problems expressible as \ac{LTL} formulae in the recurrence class \citep{manna1990hierarchy}. This algorithm, which we consider our primary contribution, is built on top of two technical novelties that address distinct challenges associated with \ac{LTL} problems. First, to overcome the poor scalability of sparse rewards, we encode automaton states with their associated goals, enabling the use of goal-conditioned techniques such as \ac{HER} \cite{andrychowicz2017her} to densify reward signals. Second, inspired by \ac{HJ} reachability analysis \cite{fisac2019bridging, hsu2021safety}, we employ a minimum-safety-based product \ac{CMDP} formulation that enforces compliance with arbitrary \ac{LTL} safety constraints at optimality. We demonstrate these technical contributions are necessary in enabling our \ac{ACQL} algorithm to significantly outperform other algorithms for solving \ac{LTL} tasks. Additionally, we demonstrate its effectiveness in learning real-world-deployable policies for a 6-DOF robot arm operating in a storage cabinet environment.

\section{Related Work}






The use of \ac{LTL} specifications with \ac{RL} algorithms has been a popular topic in recent years \citep{li2017reinforcement, li2019formal, hasanbeig2020deep, vaezipoor2021ltl2action,  araki2021logicaloptionsframework, tasse2022skill, yalcinkaya2023automata, xiong2024dscrl, xu2024generalization, yalcinkaya2024compositional, yalcinkaya2025automataconditionedrl, jackermeier2025deepltl, shah2025ltlconstrained}. One approach to this topic has been directly defining non-Markovian reward functions based on overall task satisfaction, but this only supports \ac{LTL} formulae that are satisfiable over finite prefixes \citep{li2017reinforcement, li2018policysearch, aksaray2016qlearningrobustsatisfactionsignal, balakrishnan2019structured}. In restricted cases where task satisfaction is differentiable with respect to a policy's actions, it is also possible to directly train sequence models that satisfy the specification \cite{xiong2024dscrl}. The most popular and general methods, however, convert \ac{LTL} tasks into automata to use in conjunction with standard deep \ac{RL} techniques for \acp{MDP} \cite{camacho2019ltl, quint2021formal, li2018automata, li2019formal, tasse2020booleantaskalgebrareinforcement, jothimurugan2021compositional, hasanbeig2019logicallyconstrainedreinforcementlearning, hasanbeig2020cautious, hasanbeig2020deep, yalcinkaya2023automata, xu2024generalization, yalcinkaya2024compositional, yalcinkaya2025automataconditionedrl, jackermeier2025deepltl, shah2025ltlconstrained}. Within these works, there are two subcategories including methods for learning policies that can perform well across a set of multiple \ac{LTL} tasks \cite{vaezipoor2021ltl2action, yalcinkaya2023automata, yalcinkaya2024compositional, jackermeier2025deepltl} and methods that focus on optimally satisfying a single \ac{LTL} task \cite{toroicarte2018RewardMachines, camacho2019ltl, quint2021formal, li2019formal, hasanbeig2019logicallyconstrainedreinforcementlearning, hasanbeig2020cautious, hasanbeig2020deep, toroicarteRewardMachines2022, shah2025ltlconstrained}.We specifically aim to improve on the scalability of these latter methods in high-dimensional environments. Of these techniques, we specifically highlight \acp{RM} \cite{toroicarteRewardMachines2022} as a framework with similar generality to our method and the first baseline we compare against.

Our work utilizes techniques from GCRL~\cite{kaelbling1993learning, schaul2015universal} to accelerate learning for \ac{LTL} tasks. Other methods have taken a related approach by learning or re-using goal-conditioned policies that are hierarchically guided by \ac{LTL} expressions to enable zero-shot generalization to new tasks~\citep{qiu2023instructing, yalcinkaya2023automata, jackermeier2025deepltl, yalcinkaya2025automataconditionedrl}. Using discrete skills as opposed to a goal-conditioned policy has also been explored in \cite{tasse2022skill, araki2021logicaloptionsframework}. All these approaches require the user to train a different skill for every automaton edge or a different goal-conditioned policy for every safety constraint that appears in the task, except for limited classes of safety constraints. 
This is impractical for most real-world tasks and, in particular, when safety constraints are changing throughout a task. Conversely, our method learns a single goal-conditioned policy for all goals in a particular \ac{LTL} expression while still accounting for arbitrary, potentially non-stationary safety constraints. While \ac{ACQL} has been developed for online \ac{RL} rather than zero-shot generalization, there is a significant overlap in the tasks that both can ultimately solve. Therefore, we include the \ac{LOF} \citep{araki2021logicaloptionsframework} as a representative for these methods in our experiments. 




Our work is also focused on solving \ac{LTL} tasks that feature safety constraints. Safety constraints are often indirectly addressed in the methods discussed above. They primarily focus on making unsafe actions suboptimal by either terminating rollouts \cite{toroicarte2018RewardMachines, camacho2019ltl, toroicarteRewardMachines2022, hasanbeig2019logicallyconstrainedreinforcementlearning} or applying negative reward shaping \cite{araki2021logicaloptionsframework, quint2021formal, vaezipoor2021ltl2action} when an unsafe automaton transition occurs. Other approaches based on formal methods, such as shielding \cite{alshiekh2018shielding, konighofer2023online} and runtime model-checking \cite{desai2017combining}, can more robustly enforce safety requirements but typically rely on additional assumptions such as access to a discrete abstraction of the environment or known system dynamics. In contrast, techniques from the Safe \ac{RL} literature can still robustly enforce stationary safety constraints in complex environments without such assumptions \cite{altman1999constrained, yu2022reachabilityconstrainedreinforcementlearning, kalweit2020deepconstrainedqlearning, stooke2020responsivesafetyreinforcementlearning, ha2021saclagrangian, achiam2017cpo, liu2020ipo}. Our approach builds on these Safe \ac{RL} methods to address non-stationary safety constraints induced by \ac{LTL} specifications.



\section{Preliminaries}

\paragraph{Constrained Reinforcement Learning}

\ac{RL} problems with constraints are typically modeled as discounted \acp{CMDP}, which are defined by a tuple $(\mathcal{S}, \mathcal{A}, \mathcal{T}, d_0, r, c, \mathcal{L}, \gamma)$. This tuple consists of a state space $\mathcal{S}$, an action space $\mathcal{A}$, a transition function $\mathcal{T} : \mathcal{S} \times \mathcal{A} \rightarrow P(\mathcal{S})$ (where $P(\mathcal{S})$ represents the space of probability measures over $\mathcal{S}$), an initial state distribution $d_0 \in P(\mathcal{S})$, reward and constraint functions $r, \; c : \mathcal{S} \times \mathcal{A} \rightarrow \mathbb{R}$, a discount factor $\gamma$, and a limit for constraint violation $\mathcal{L} \in \mathbb{R}$ \cite{altman1999constrained}. Given a \ac{CMDP}, the standard objective is to find the stationary policy $\pi:\mathcal{S} \to P(\mathcal{A})$ satisfying the constrained maximization
\begin{equation}
    \label{eqn:normal-cmdp-objective}
    \max_{\pi} \; J_r(\pi) \quad \textnormal{s.t.} \quad J_c(\pi) < \mathcal{L},
\end{equation}
    where $J_{r}(\pi) = \expectation_{\tau \sim \pi} \left [ \sum_{t=0}^{\infty} \gamma^t r(s_t, a_t) \right ]$ and $J_{c}(\pi) = \expectation_{\tau \sim \pi} \left [ \sum_{t=0}^{\infty} \gamma^t c(s_t, a_t) \right ]$ are respectively the expected total discounted return and cost of the policy $\pi$ over trajectories $\tau = (s_0, a_0, s_1, a_1, \dots)$ induced by $\pi$ through interactions with the environment of the \ac{CMDP}. We denote the state value function of $\pi$ by $V^r_{\pi}(s) = \mathbb{E}_{\tau}[\sum_{t=0}^{\infty}\gamma^t r(s_t,a_t) | s_0=s]$ and the state-action value function by $Q^r_{\pi}(s,a) = \mathbb{E}_{\tau}[\sum_{t=0}^{\infty}\gamma^t r(s_t,a_t)|s_0=s, a_0=a]$.

\paragraph{Temporal Logic}
\label{sec:TL}

\ac{LTL} \cite{pnueli1977ltl} is an extension of propositional logic for reasoning about systems over time. \ac{LTL} formulae $\phi \in \Phi$ consist of atomic propositions $p$ from a set $AP$, the standard boolean operators ``not'' ($\neg$), ``and'' ($\wedge$), and ``or'' ($\vee$), and temporal operators that reference the value of propositions in the future. These operators are ``next'' ($\circ$), ``eventually'' ($\lozenge$), and ``always'' ($\square$). We follow the definitions for these operators given in \cite{baier2008principles}. In our algorithm, we use \ac{STL}, which further extends \ac{LTL} with quantitative semantics and requires that every atomic proposition $p \in AP$ is defined as a real-valued function of the system state. The quantitative semantics is defined by a function $\rho : \mathcal{S}^\omega \times \Phi \rightarrow \mathbb{R}$ that produces a robustness value 
representing how much a sequence of states $\omega \in \mathcal{S}^{\omega}$ satisfies or violates the property specified by an \ac{STL} formula $\phi \in \Phi$. Our exact implementation of these quantitative semantics is based on~\cite{dawson2022robust_stl_planning}. Every temporal logic constraint can be expressed as the conjunction of a ``safety'' and ``liveness'' constraint \cite{alpern1985liveness}, where a safety constraint is informally defined as a requirement that something must never happen for the task to succeed, and a liveness constraint is defined as a requirement that something must happen for the task to succeed. We take advantage of this dichotomy in our method.

\paragraph{Automata}
\label{section:automata-preliminaries}

The robotics tasks discussed so far can be expressed more specifically as \ac{STL} formulae in the recurrence class, as defined in \cite{manna1990hierarchy}, and all such formulae can be translated into abstract machines called \acp{DBA} \cite{baier2008principles}. A \ac{DBA} is formally defined by the tuple $A = (\Sigma, \mathcal{Q}, \delta, q_0, F)$, consisting of an alphabet $\Sigma$, a set of internal states $\mathcal{Q}$, a transition function $\delta : \mathcal{Q} \times \Sigma \rightarrow \mathcal{Q}$, an initial state $q_0 \in \mathcal{Q}$, and a set of accepting states $F \subseteq \mathcal{Q}$. An automaton is used to process an infinite sequence of arbitrary symbols from the alphabet $\Sigma$ and determine whether the sequence satisfies or does not satisfy the original logical expression. In our setting, a symbol processed by the automaton is a subset of atomic propositions $l \in 2^{AP}$ that are true in some \ac{MDP} state $s \in \mathcal{S}$. This symbol $l$ is referred to as the state's labeling and the mapping of states to their labeling is denoted by a labeling function $L : \mathcal{S} \to 2^{AP}$. The automaton $A$ is always in some state $q \in \mathcal{Q}$, and starts in the state $q_0$. Each element $\sigma \in \Sigma$ of an input sequence causes some change in the automaton's internal state according to the transition function $\delta$. For convenience, we will refer to edges in the automaton with transition predicates. A transition predicate is a propositional formula that holds for all $\sigma \in \Sigma$ which induce a transition between two states $q_i, q_j \in \mathcal{Q}$ according to $\delta$ \cite{alpern1987Recognizing}. By processing each state while an agent interacts with an \ac{MDP}, the internal automaton state provides information to the agent about its current progress in the task. This is the fundamental idea behind a Product MDP, which combines any \ac{MDP} with a deterministic automaton and is a standard technique in model checking for probabilistic systems \cite{baier2008principles}. A complete definition for a product \ac{MDP} in an \ac{RL} context can be found in previous work such as \cite{hasanbeig2020deep, toroicarteRewardMachines2022}.

\section{Method}

\begin{figure*} 
\centering



\begin{tikzpicture}

	\node (image) at (0,0) {
	    \includegraphics[width=\linewidth]{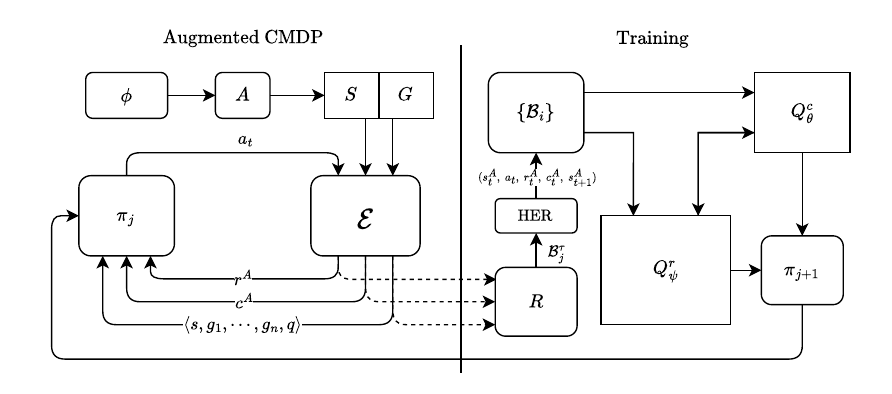}
    };

    \node (normal-annot) at (4.4,0.1) {
        \eqref{eqn:normal-bellman-loss-and-target}
    };

    \node (safety-annot) at (4.4,1.8) {
        \eqref{eqn:safety-bellman-target}
    };

\end{tikzpicture}

\caption{The \ac{ACQL} algorithm relies on a novel augmented formulation of \acp{CMDP} (left). An input task specification $\phi$ is converted into the \ac{DBA} $A$, from which safety constraints and subgoals are collected into mappings $S$ and $G$ respectively. The learning agent receives subgoals $g_1, \cdots, g_n$ at every stage in the task from $G$ and safety constraint feedback in $c^A$ from $S$. Trajectories induced by the policy $\pi_j$ are collected into a replay buffer $R$, from which batches $\mathcal{B}^{\tau}_j$ are sampled and modified using \ac{HER} \cite{andrychowicz2017her}. From these modified trajectories, mini-batches $\mathcal{B}_i$ of transitions $(s^A_t, a_t, r^A_t, c^A_t, s^A_{t+1})$ are used to compute the targets $y_t^r$ in \eqref{eqn:normal-bellman-loss-and-target} and $y_t^c$ in \eqref{eqn:safety-bellman-target} for training models of the state-action value function and safety function, $Q^r_\theta$ and $Q^c_\psi$, from which an updated policy $\pi_{j+1}$ is derived.}

\label{fig:pipeline}

\end{figure*}

This section introduces our \ac{ACQL} algorithm along with a novel augmented product \ac{CMDP} on which it is based. An overview is provided in Figure~\ref{fig:pipeline}. First, we detail the construction of the augmented product \ac{CMDP} derived from a task specification. Observations in this \ac{CMDP} include the set of subgoals associated with the current automaton state $q$, as described in Section~\ref{section:automata-preliminaries}. This \ac{CMDP} also uses safety conditions obtained from the task automaton to provide separate constraint feedback that the learning agent uses to determine safe actions. Unlike the standard \ac{CMDP} constraint in \eqref{eqn:normal-cmdp-objective}, we argue for the use of a constraint on the minimum safety that is easier to learn. We finally present an overview and brief analysis of \ac{ACQL}, which is tailored to utilize the information provided by our augmented product \ac{CMDP} to scale beyond existing algorithms for learning in product \acp{MDP}.

\subsection{Augmented Product \ac{CMDP} Formulation}

\paragraph{Obtaining the Automaton, Safety, and Liveness Constraints}
\label{section:obtaining-a}
\label{section:obtaining-s}
\label{section:obtaining-g}

To construct the augmented product \ac{CMDP} used by our algorithm, we begin by translating the input \ac{STL} task specification $\phi$ into a \ac{DBA} $A = (\Sigma, \mathcal{Q}, \delta, q_0, F)$ using the SPOT library \citep{duret22cav}. As described in Section~\ref{section:automata-preliminaries}, this automaton represents the structure of the task. To better guide learning, we extract from this automaton additional structure that captures the task's safety and liveness requirements. Intuitively, these represent conditions that must \emph{always} hold to avoid failure (safety) and conditions that must \emph{eventually} hold to make progress toward task completion (liveness) \citep{alpern1985liveness, alpern1987Recognizing}. This information is summarized in two mappings: a safety condition mapping $S : \mathcal{Q} \rightarrow \Phi$, which assigns to each automaton state a proposition that must remain true, and a liveness condition mapping $O : \mathcal{Q} \rightarrow \Phi$, which assigns to each automaton state a proposition that must be eventually satisfied to proceed in the task. Based on $O$, we define $G : \mathcal{Q} \rightarrow \mathcal{G}^+$ to explicitly link automaton states to subgoals, where $\mathcal{G}^+ = \bigcup_{i=1}^n \mathcal{G}^i$ represents the full space of possible subgoal lists up to a task-dependent length $n$ and $\mathcal{G}^i$ is the $i$-fold cartesian product of $\mathcal{G}$ with itself.


First, to obtain safety constraints, we determine the non-accepting sink-components of the automaton $A$; i.e., the automaton states $Q \setminus F$ from which there is no path to any state in $F$.  For each automaton state $q$, we then examine the outgoing transitions from $q$ that lead into these sink components. The predicates guarding these transitions describe conditions under which a transition into an unsafe state would occur, so negating these predicates provides the conditions that must hold in order to stay safe while in state $q$. We can then define $S(q)$ as the conjunction of all such negated predicates for state $q$.

Second, to obtain liveness constraints, we ignore the above transitions for safety conditions and 
examine the remaining predicates guarding outgoing transitions from each state $q$. These can only specify the conditions necessary to make progress in the task. Therefore, we can take the disjunction of these transition predicates to obtain
the formula $O(q)$ that must be satisfied in order to transition beyond $q$. Following prior work on learning \ac{LTL} tasks~\cite{araki2021logicaloptionsframework, qiu2023instructing}, we assume that a subset of the atomic propositions, $AP_{\text{subgoal}} \subseteq AP$, represent subgoal propositions and are parameterized by values $g$ within a subgoal space $\mathcal{G} \subseteq \mathcal{S}$. Note that we consider only a subset since not all the propositions in our modified \ac{DBA} are necessarily related to achieving subgoals. Using this fact, we filter the predicates in $O(q)$ to retain only those associated with these subgoal propositions, which in turn provides a list of subgoals $G(q) = g_1, \dots, g_n$ that are relevant to the logical formula given by $O(q)$. An illustrative example for these two steps is provided in the appendix.

\paragraph{Defining the Augmented Product \ac{CMDP}} 
\label{section:making-m}
Now, let $\mathcal{M}^{A} = (\mathcal{S}^{A}, \mathcal{A}^A, \mathcal{T}^{A}, d_0^{A}, r^{A}, c^{A}, \gamma, \mathcal{L})$ be a new \ac{CMDP} formed by augmenting the original \ac{MDP} $\mathcal{M} = (\mathcal{S}, \mathcal{A}, \mathcal{T}, d_0, r, \gamma)$ with the \ac{DBA} $A = (\Sigma, \mathcal{Q}, \delta, q_0, F)$. The new state space $\mathcal{S}^A = \mathcal{S} \times \mathcal{G}^+ \times \mathcal{Q}$ is the Cartesian product of the original state space $\mathcal{S}$, the space for every possible list of subgoals $\mathcal{G}^+$, and the set of automaton states $\mathcal{Q}$. A state $s^A \in \mathcal{S}^A$ can be written as $\langle s, g^+, q \rangle$, where $s \in \mathcal{S}$, $g^+ \in \mathcal{G}^+$, and $q \in \mathcal{Q}$ are the constituent \ac{MDP} state, goal-list and automaton state of $s^A$. The action space $\mathcal{A}^A = \mathcal{A}$ is unmodified. The new transition dynamics $\mathcal{T}^A$ are defined so that a transition to $\langle s', g^+{'}, q' \rangle \in \mathcal{S}^A$ from $\langle s, g^+, q \rangle \in \mathcal{S}^A$ is impossible if the automaton does not support a transition from $q$ to $q'$ when entering the state $s'$; i.e.,
\begin{align*}
    \mathcal{T}^A&(\langle s', g^{+}{'}, q' \rangle | \langle s, g^+, q \rangle, a) = \begin{cases}
        \mathcal{T}(s'|s, a) & \text{if}\ q' = \delta(q, L(s')), g^+{'} = G(q'), \\
        0 & \text{otherwise}.
    \end{cases}
\end{align*}
Likewise, the initial state distribution $d_0^A(\langle s, g^+, q \rangle) = d_0(s)$ if $q = q_0$ and $g^+ = G(q_0)$, and is zero otherwise. The reward function $r^A(\langle s, g^+, q \rangle) = \mathbbm{1}_{F}(q)$, where $\mathbbm{1}(\cdot)$ is an indicator function, is defined to provide a sparse reward of 1 when the task is finished and the agent is near the goal associated with any accepting state of the automaton. Reward sparsity was a central shortcoming of prior work, often resulting in poor performance in realistic scenarios. However, our augmented product \ac{CMDP} specifically includes the subgoal list $g^+$ to facilitate the use of modern \ac{GCRL} algorithms to mitigate this sparsity in many relevant tasks. In particular, \ac{ACQL} retroactively assigns rewards based on achieved subgoals using \ac{HER}, and the benefit of this strategy for \ac{LTL}-specified tasks is validated in our ablations (Section~\ref{section:ablations}). Lastly, we define the constraint function $c^A$ along with the constrained objective used with our \ac{CMDP} formulation.


\paragraph{Minimum \ac{LTL} Safety Constraint} 
Although the sum-of-costs formulation in \eqref{eqn:normal-cmdp-objective} conveniently leverages the standard Bellman equation, it presents practical challenges in learning. Specifically, predicting the cumulative sum of future costs requires accurately modeling long-term dependencies, which increases variance and slows down convergence. This can be appropriate for maximizing reward but unnecessarily complicated for safety. When dealing with safety. and in particular hard-safety \cite{wang2023hardconstraints}, we are interesting in knowing whether the generated trajectory becomes unsafe at least once. Hence, this can be reformulated as a classification task. This approach bypasses the need for accurate regression by directly learning the decision boundary for state-action pairs leading to safety violations. To achieve this, we define a constraint function $c^A$ such that it takes negative values if a safety constraint is violated and positive values otherwise, and we apply a constraint based on the minimum over all future values of $c^A(s^A_t, a_t)$. Then, we simply learn to classify this minimum value as being either positive or negative. This formulation allows us to determine safety violations in a single step, avoiding the complexity of cumulative cost prediction. A major advantage of this formulation is that it removes the need for manual tuning of the violation limit $\mathcal{L} \in \mathbb{R}$, which depends on the range of possible costs. Instead, by directly classifying actions as leading to a safety violation, we reduce the range of $\mathcal{L}$ to $[-1,1]$. This normalization ensures robustness across tasks and allows for a simple, task-agnostic choice of $\mathcal{L} = 0$. Directly defining $c^A(\langle s, g^+, q \rangle, a) = \rho(s, S(q))$ using the robustness function for $S(q)$ (see Section~\ref{sec:TL}) satisfies this formulation and corresponds to the safety constraints for any \ac{STL}-specified task. 

Now, we can define our new objective as finding the optimal stationary policy $\pi^* : \mathcal{S} \to P(\mathcal{A})$ that maximizes the expected total discounted return, as defined in \eqref{eqn:normal-cmdp-objective}, while keeping the expected minimum safety above the safety limit $\mathcal{L}$:
\begin{equation}
    \label{eqn:min-cost-cmdp}
    \expectation_{\tau \sim \pi} \left [ \min_{t \in [0, \infty]} c^A(s_t, a_t) \right ] > \mathcal{L},
\end{equation}
where $\tau = (s^A_0,a_1,s^A_2,a_2,\dots)$ is a trajectory within our augmented \ac{CMDP} induced by the policy $\pi$. We denote this expected minimum safety for a policy $\pi$ by its state-action safety function $Q^c_\pi(s, a) = \mathbb{E}_{\tau \sim \pi}[\min_{t=0}^{\infty} c^A(s^A_t,a_t)|s^A_0=s, a_0=a]$. 


\subsection{\acf{ACQL}}



\begin{algorithm}[t]
\caption{Automaton Constrained Q-Learning} \label{alg:achql}
\begin{algorithmic}[1]
    \Require An MDP $\mathcal{M} = (\mathcal{S}, \mathcal{A}, \mathcal{T}, d_0, r, \gamma)$, an \ac{STL} specification $\phi \in \Phi$, a safety limit $\mathcal{L} \in [-1, 1]$, a learning rate $\alpha$, an interpolation factor $\lambda$
    \State $A \leftarrow \Call{Translate}\phi$
    \State $S, G \leftarrow \Call{Partition}A$
    \State $\mathcal{M}^A \leftarrow (\mathcal{S}^A, \mathcal{A}, \mathcal{T}^A, d_0^A, r^A, c^A, \gamma, \mathcal{L})$ 
    \State $Q^c_{\theta}, Q^r_{\psi} \leftarrow \Call{MakeNetworks}{\mathcal{S^A}, \mathcal{A}}$
    \State $\bar{\theta} \leftarrow \theta, \bar{\psi} \leftarrow \psi$
    \State $R \leftarrow \Call{MakeReplayBuffer}{\mathcal{M}^A}$
    \For{$j = 1, \ldots, N$}
        \State $\gamma_c \leftarrow \Call{SafetyGammaScheduler}{j}$
        \State $\pi_j(s^A) \leftarrow \argmax_{a : Q^c_{\theta}(s^A, a) > \mathcal{L}}Q^r_{\psi}(s^A, a)$
        \State $\tau \leftarrow \Call{GetTrajectory}{\mathcal{M}^A, \pi_j}$
        \State $R \leftarrow R \cup \tau$.
        \State $\mathcal{B}^{\tau}_j \sim R$
        \State $\mathcal{B}^{\tau}_j \leftarrow \Call{Relabel}{\mathcal{B}^{\tau}_j}$
        \For{$i = 1, \ldots, M$}
            \State $\mathcal{B}_i \leftarrow \{ (s^A_t, a_t, r^A_t, c^A_t, s^A_{t+1}) \} \sim \mathcal{B}^{\tau}_j$
            \State $\theta \leftarrow \theta - \alpha \nabla_{\theta} L^c_{j,i}(\theta)$
            \State $\psi \leftarrow \psi - \alpha \nabla_{\psi} L^r_{j,i}(\psi)$
            \State $\bar{\theta} \leftarrow (1 - \lambda) \bar{\theta} + \lambda \theta, \bar{\psi} \leftarrow (1 - \lambda)\bar{\psi} + \lambda \psi$
            
        \EndFor
    \EndFor
\end{algorithmic}
\end{algorithm}

\paragraph{Overview}
\label{section:acql-overview}
In what follows, we provide an overview of the \ac{ACQL} algorithm, whose pseudocode is shown in Algorithm~\ref{alg:achql}. First, the input \ac{STL} specification $\phi$ is translated into an automaton $A$ (Line 1) that is partitioned to produce the mappings $S$ and $G$ (Line 2). These mappings and the automaton are combined with the input \ac{MDP} $\mathcal{M}$ to create the augmented \ac{CMDP} in Line 3. Using the augmented \ac{CMDP}, \ac{ACQL} can proceed to learn the optimal policy 
\[  \pi^*(s^A) = \argmax_{a \; : \; Q^{c*}(s^A, a) > \mathcal{L}} Q^{r*}(s^A, a), 
\]
by learning the optimal state-action value function $Q^{r*} = Q^r_{\pi^*}$ and the optimal state-action safety function $Q^{c*} = Q^c_{\pi^*}$. In order to learn these optimal functions, we define the models $Q^r_\psi : \mathcal{S}^A \times \mathcal{A} \rightarrow \mathbb{R}$ and $Q^c_\theta : \mathcal{S}^A \times \mathcal{A} \rightarrow [-1, 1]$ parameterized by $\psi$ and $\theta$ (Line 4). Their initial parameters are copied to initialize the target parameters $\bar{\theta}$ and $\bar{\psi}$ in Line 5, and an empty replay buffer is initialized in Line 6.

The algorithm proceeds to iterate for $N$ epochs indexed by $j$. At each epoch, we obtain a value for the safety discount factor $\gamma_c$ in Equation~\eqref{eqn:safety-bellman-target}, which asymptotically approaches $1.0$ as training progresses (Line 8). The policy $\pi_j$ for the epoch is defined to select the most-rewarding action according to $Q^{r}_{\psi}$ constrained by $Q^c_{\theta}$ (Line 9). A trajectory $\tau$ is collected according to an epsilon-greedy version of this policy (Line 10), and this trajectory is added to the replay buffer (Line 11). From this replay buffer, a batch of trajectories $\mathcal{B}^{\tau}_j$ is sampled every epoch (Line 12). The inclusion of subgoals in the states $s^A_t$ of the sampled mini-batch $\mathcal{B}_i$ allows us to further accelerate learning using relabeling techniques such as \ac{HER} \cite{andrychowicz2017her}, which allows $Q^r_\theta$ to improve from failed attempts at progressing through the task. Mini-batches of transitions $\mathcal{B}_i$ are subsequently sampled from the relabeled trajectories (Line 15) and used to update $Q^r_\psi$ for minimizing the loss
\begin{equation}
    \label{eqn:normal-bellman-loss-and-target}
    L^r_{j,i}(\psi) = \expectation_{(s^A_t, a_t) \sim \mathcal{B}_i} \left [ \left ( y^r_t - Q^r_{\psi}(s^A_t, a_t) \right )^2 \right ], \ \ \text{with} \ \
    y^r_t = r^A_t + \gamma Q^r_{\bar{\psi}}(s^A_{t+1}, \pi_{j}(s^A_{t+1})).
\end{equation}
Similarly, $Q^c_\theta$ is trained with the target
\begin{align}
    \label{eqn:safety-bellman-target}
    y^c_t =& \gamma_c \min \{ c^A(s^A_t, a_t), \; Q^{c}_{\bar{\theta}}(s^A_{t+1}, \pi_{j}(s^A_{t+1})) \} + (1 - \gamma_c) \; c^A(s^A_t, a_t),
\end{align}
which is derived from the Bellman principle of optimality for expected minimum cost objectives given in \cite{fisac2019bridging}. The targets, $y^r$ and $y^c$, are computed using target parameters, $\bar{\psi}$ and $\bar{\theta}$, which are updated towards their corresponding main parameters each step with an interpolation factor $\lambda$ (Line 18). Crucially, we schedule the discount factor $\gamma_c$ throughout training to asymptotically approach $1.0$, which is necessary for convergence to $Q^{c*}$. Additional details for the subroutines used in Algorithm~\ref{alg:achql} are provided in the appendix.

\paragraph{Analysis}
\label{section:analysis}
Under mild assumptions, \ac{ACQL} is guaranteed to asymptotically converge to the optimal solution. We summarize this in the following proposition:
\begin{proposition}
    \label{proposition:convergence}
    Let $\mathcal{M^A}$ be an augmented \ac{CMDP} with $|\mathcal{S}^A| < \infty$, $|\mathcal{A}| < \infty$, and $\gamma \in [0, 1)$, and let $Q^c_n$ and $Q^r_n$ be models for the state-action safety and value functions indexed by $n$. Assume they are updated using Robbins-Monro step sizes $a(n)$ and $b(n)$, respectively, with $b(n) \in o(a(n))$ according to \eqref{eqn:normal-bellman-loss-and-target} and \eqref{eqn:safety-bellman-target}. Assume that $\gamma_{c_n}$ is also updated with step sizes $c(n)$ such that $\gamma_{c_n} \rightarrow 1$ and $c(n) \in o(b(n))$. Then $Q^c_n$ and $Q^r_n$ converge to $Q^{c*}$ and $Q^{r*}$ almost surely as $n \rightarrow \infty$.
    \begin{proofsketch}
        The step sizes $a(n)$, $b(n)$, and $c(n)$ create a stochastic approximation algorithm on three timescales \cite{borkar2008stochasticapproximation}. Based on the contraction property of \eqref{eqn:safety-bellman-target}, the fastest updating $Q^c_n$ asymptotically tracks the correct safety function for the policy determined by $Q^r_n$ and $\gamma_{c_n}$. On the slower timescale ensured by the definition of $b(n)$ and $a(n)$, $Q^r_n$ converges to the fixed point as determined by the relatively static $\gamma_{c_n}$, due to the contraction property of \eqref{eqn:normal-bellman-loss-and-target}. Finally, as $\gamma_c$ converges on the slowest timescale, the fixed points for the two faster timescales approach $Q^{r*}$ and $Q^{c*}$ at $\gamma_c = 1$.
    \end{proofsketch}
\end{proposition}
In Proposition~\ref{proposition:convergence}, the notation $b(n) \in o(a(n))$ means that $b(n)/a(n) \to 0$ asymptotically. The assumption of a finite \ac{MDP} is standard in the literature, but relaxing this assumption is possible by applying fixed point theorems for infinite dimensional spaces \citep{granas2003fixed}. After convergence, the final policy will behave optimally for the reward $r$ and (if possible) never incur any cost, thereby respecting the \ac{LTL} safety constraint. A complete description of our algorithm, model implementation details, and a full proof of convergence, with reference to similar arguments used in prior work \cite{ma2022jointsynthesissafetycertificate, chow2017riskconstrainedreinforcementlearningpercentile, yu2022reachabilityconstrainedreinforcementlearning}, are provided in the appendix.



\section{Experiments}
\label{section:experiments}

\newcommand{\sequence}{$\lozenge(g_1 \wedge \circ(\lozenge g_2))$}
\newcommand{\branch}{$\lozenge g_1 \wedge \lozenge g_2$}
\newcommand{\obligation}{$\lozenge g_1 \wedge \square \neg o_1$}
\newcommand{\until}{$\neg o_1 \mathcal{U} g_1 \wedge \circ \lozenge g_2$}
\newcommand{\looptask}{$\square \lozenge (g_1 \wedge \circ \lozenge g_2) \wedge \square \neg o_1$}

\afterpage{
\begin{figure*}
    \centering
  \subfloat[PointMass Field]{%
       \includegraphics[width=0.21\linewidth]{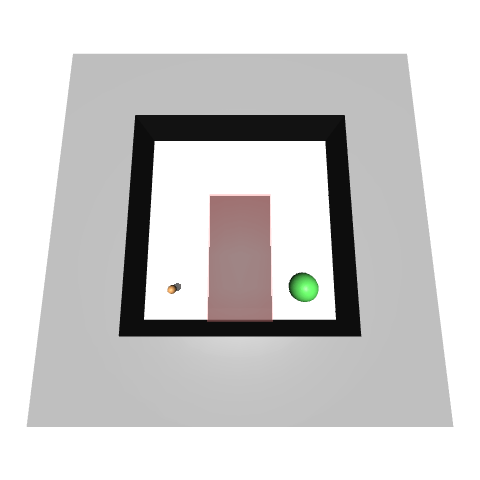}
       \label{fig:environments:point_maze}
    }
  \hspace{0.005\linewidth}   
  \subfloat[Quadcopter Room]{
        \includegraphics[width=0.21\linewidth]{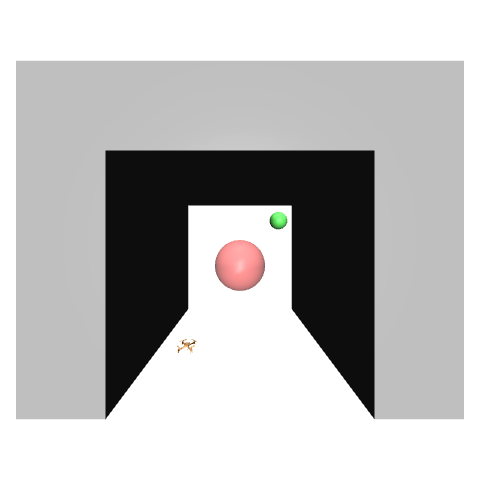}
        \label{fig:environments:point_maze_3d}
    }
  \subfloat[Ant Field]{%
        \includegraphics[width=0.21\linewidth]{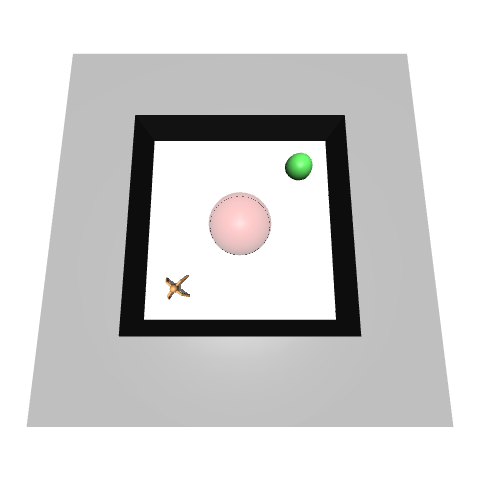}
        \label{fig:environments:ant_maze}
    }
  \subfloat[UR5e Shelf]{%
        \includegraphics[width=0.21\linewidth]{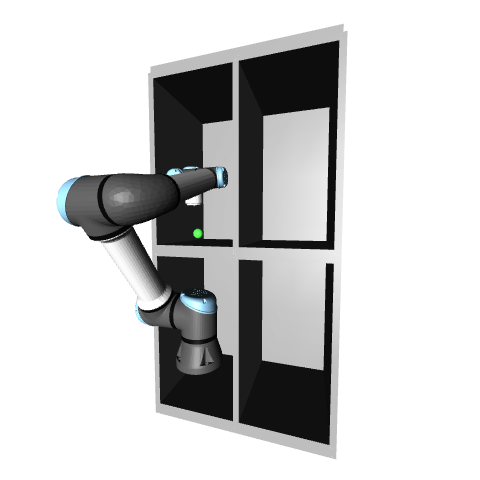}
        \label{fig:environments:ur5e_shelf}
    }
  \caption{We conducted experiments in simulated environments for navigation tasks using a point mass, quadcopter, and ant quadruped. We also trained policies for end-effector control of a UR5e manipulator in a simulated shelf-environment for our real-world experiments.}
  
  
  \label{fig:environments}
\end{figure*}
\begin{table*}[h]




\centering
    \caption{Results from policies trained for five task types in three different environments with 5 million environment interactions and 5 seeds. We report mean and one standard deviation of total reward and success rate in 16 evaluation episodes lasting 1000 steps over seeds for policies at the end of training.}
    \label{tab:task-experiments}
    \tiny
    \setlength{\tabcolsep}{4.0pt}
    \begin{tabular}{lccccccc}
        \toprule
        \multicolumn{2}{c}{} & \multicolumn{2}{c}{{ACQL (Ours)}} & \multicolumn{2}{c}{{LOF} \cite{araki2021logicaloptionsframework}} & \multicolumn{2}{c}{{CRM-RS} \cite{toroicarteRewardMachines2022}} \\
        
        \cmidrule{3-8}
        \multicolumn{1}{c}{Robot} & Task & Reward $\uparrow$ & $S.R. (\%)$ $\uparrow$ & Reward $\uparrow$ & $S.R. (\%)$ $\uparrow$ & Reward $\uparrow$ & $S.R. (\%)$ $\uparrow$ \\
        \midrule
        \multirow{5}{*}{P.M.} &
        {\sequence} &
        \boldsymbol{$829.6 \pm 9.6$} & \boldsymbol{$83.8 \pm 26.4$} &
        $11.0 \pm 3.1$ & $98.8 \pm 2.8$ &
        $91.2 \pm 4.3$ & $0.0 \pm 0.0$ \\
        & {\branch}  &
        \boldsymbol{$841.1 \pm 54.1$} & \boldsymbol{$98.8 \pm 2.8$} &
        $470.7 \pm 397.5$ & $100.0 \pm 0.0$ &
        $420.6 \pm 259.2$ & $61.3 \pm 38.6$ \\
        & {\obligation} &
        \boldsymbol{$858.7 \pm 3.2$} & \boldsymbol{$100.0 \pm 0.0$} &
        $0.0 \pm 0.0$ & $0.0 \pm 0.0$ &
        $0.0 \pm 0.0$ & $3.8 \pm 8.4$ \\
        & {\until} &
        \boldsymbol{$525.8 \pm 480.0$} & \boldsymbol{$60.0 \pm 54.8$} &
        $0.0 \pm 0.0$ & $0.0 \pm 0.0$ &
        $0.0 \pm 0.0$ & $2.5 \pm 5.6$ \\
        & {\looptask} &
        \boldsymbol{$2.2 \pm 2.0$} & \boldsymbol{$62.5 \pm 41.8$} &
        $0.0 \pm 0.0$ & $0.0 \pm 0.0$ &
        $0.0 \pm 0.0$ & $0.0 \pm 0.0$ \\
        \midrule
        \multirow{5}{*}{Q.C.} &
        {\sequence}     &
        \boldsymbol{$813.5 \pm 9.7$} & \boldsymbol{$100.0 \pm 0.0$} & 
        $32.0 \pm 6.3$ & $98.8 \pm 2.8$ &
        $0.0 \pm 0.0$ & $0.0 \pm 0.0$ \\
        & {\branch}  &
        \boldsymbol{$752.6 \pm 157.0$} & \boldsymbol{$92.5 \pm 16.8$} & 
        $161.7 \pm 287.6$ & $92.5 \pm 16.8$ &
        $0.0 \pm 0.0$ & $0.0 \pm 0.0$ \\
        & {\obligation} &
        \boldsymbol{$822.7 \pm 107.4$} & \boldsymbol{$95.0 \pm 11.2$} & 
        $3.5 \pm 5.9$ & $15.0 \pm 20.5$ &
        $4.1 \pm 9.2$ & $2.5 \pm 5.6$ \\
        & {\until} &
        \boldsymbol{$726.6 \pm 88.6$} & \boldsymbol{$97.5 \pm 3.4$} & 
        $0.0 \pm 0.0$ & $0.0 \pm 0.0$ &
        $0.0 \pm 0.0$ & $0.0 \pm 0.0$ \\
        & {\looptask} &
        \boldsymbol{$2.4 \pm 1.2$} & \boldsymbol{$82.5 \pm 35.0$} &
        $0.0 \pm 0.0$ & $0.0 \pm 0.0$ &
        $0.0 \pm 0.0$ & $0.0 \pm 0.0$ \\
        \midrule
        \multirow{5}{*}{Ant} &
        {\sequence} &
        \boldsymbol{$555.8 \pm 39.8$} & \boldsymbol{$98.8 \pm 2.8$} &
        $53.2 \pm 29.8$ & $91.2 \pm 9.5$ &
        $5.5 \pm 12.3$ & $1.2 \pm 2.8$ \\
        & {\branch}  &
        \boldsymbol{$587.2 \pm 19.3$} & \boldsymbol{$98.8 \pm 2.8$} &
        $158.0 \pm 180.6$ & $76.2 \pm 42.7$ &
        $4.6 \pm 6.4$ & $2.5 \pm 3.4$ \\
        & {\obligation} &
        \boldsymbol{$683.7 \pm 31.1$} & \boldsymbol{$85.0 \pm 3.4$} &
        $38.3 \pm 28.8$ & $8.8 \pm 7.1$ &
        $0.0 \pm 0.0$ & $3.8 \pm 3.4$ \\
        & {\until} &
        \boldsymbol{$193.4 \pm 36.2$} & \boldsymbol{$87.5 \pm 10.8$} &
        $0.0 \pm 0.0$ & $0.0 \pm 0.0$ &
        $0.0 \pm 0.0$ & $0.0 \pm 0.0$ \\
        & {\looptask} &
        \boldsymbol{$0.9 \pm 0.6$} & \boldsymbol{$77.5 \pm 12.9$} &
        $0.0 \pm 0.0$ & $0.0 \pm 0.0$ &
        $0.0 \pm 0.0$ & $0.0 \pm 0.0$ \\
        \bottomrule
    \end{tabular}
\end{table*}
}

In this section, we justify the design of \ac{ACQL} and evaluate its effectiveness. We conducted a comparative analysis of \ac{ACQL} against established baselines in \ac{RL} from \ac{LTL} specifications, including \acp{RM} \cite{toroicarte2018RewardMachines} and the \ac{LOF} \cite{araki2021logicaloptionsframework}. We demonstrate our method's real-world applicability by solving more complex \ac{LTL} tasks with a 6-DOF robot arm in a storage cabinet environment. Lastly, we performed an ablation study to clarify how our subgoal-including product \ac{CMDP}, in combination with \ac{HER} \cite{andrychowicz2017her}, and our minimum-safety constraint formulation contribute to \ac{ACQL}'s performance.

\subsection{Comparative Analysis}
\label{section:experiments-comparison}

\paragraph{Baselines}
In selecting our baselines, we chose to not compare against standard Safe \ac{RL} methods, offline policy learning methods, and multi-task \ac{LTL} methods. Safe \ac{RL} methods, without significant modification, cannot directly apply to \ac{LTL} tasks and instead only address tasks with stationary safety constraints. Offline methods assume access to a fixed dataset and are not designed for the online learning setting we address. Multi-task methods \cite{vaezipoor2021ltl2action, yalcinkaya2023automata, yalcinkaya2024compositional, jackermeier2025deepltl} take special measures to generalize across a distribution of \ac{LTL} specifications. Comparisons to these methods in a single-task context would therefore require removing significant components from them and would not yield strong conclusions regarding their relative merit. We instead compared our algorithm against two methods for online \ac{RL} with singular \ac{LTL} specifications: \ac{CRM-RS} \cite{toroicarteRewardMachines2022} and the \ac{LOF} \cite{araki2021logicaloptionsframework}.
\ac{CRM-RS} employs an \ac{RM} to define the task and dispense shaped rewards that incentivize task completion. \ac{RM} states in non-accepting sink components are treated as terminal states, implementing a rollout-terminating strategy to enforce safety constraints. The \ac{LOF} is a framework for learning hierarchical policies to satisfy \ac{LTL} tasks. It learns a separate policy, referred to as a logical option, for satisfying every subgoal-proposition in the task. When training each logical option, each step that violates the task's safety propositions receives a large reward penalty (set to $R_s = -1000$ based on their implementation) to discourage unsafe behavior. These policies are then composed with a higher-level policy obtained using value iteration over a discrete state space derived from the task automaton and all the subgoal states of the environment. Both these approaches can handle general \ac{LTL} specifications and, to our knowledge, represent the most relevant baselines to compare our algorithm against.
Furthermore, we note that \acp{RM} and the \ac{LOF} cannot be easily enhanced with high-performing \ac{GCRL} and Safe \ac{RL} techniques without adopting substantial modifications to their learning algorithms and product \ac{MDP} formulation. Our method effectively integrates these components in a single framework that enables learning from general logical objectives in complex settings, as demonstrated in the following tasks.


\vspace{-0.3cm}
\paragraph{Tasks}
We chose five distinct \ac{LTL} tasks and three different agents to facilitate a thorough comparison between our algorithm and the two baseline methods. All environment simulation was done within the Brax physics simulator \cite{brax2021github}. The agents used in our experiments are a 2D PointMass, a Quadcopter, and an 8-DOF Ant quadruped. Additional details for these environments are given in the appendix. We used open environments without physical obstacles for our experiments. Instead, obstacles were introduced solely through the task specification. This was done to demonstrate the effectiveness of our algorithm in avoiding obstacles based purely on feedback from the task specifications, without confounding effects such as physically restricted motion.
The \ac{LTL} tasks used in the evaluation include: 
(1) a two-subgoal sequential navigation task between opposite corners of the environment (\sequence),
(2) a two-subgoal branching navigation task between opposite corners of the environment (\branch), 
(3) a single-goal navigation task constrained by an unsafe region (\obligation), 
(4) a two-subgoal navigation task with a disappearing safety constraint (\until), and 
(5) an infinitely-looping navigation task with a persistent safety constraint (\looptask). The starting position, obstacle and subgoal configurations for each environment are shown in Figure~\ref{fig:environments}. 

\vspace{-0.3cm}
\paragraph{Results} The results for our three simulated environments and five task types are shown in Table~\ref{tab:task-experiments}. All results are reported for the final policy obtained after 5 million environment interactions with five different seeds. For \ac{LOF}, each individual logical option was afforded its own 5 million training steps for fairness. For each seed, we evaluated the final policy in 16 randomly initialized episodes lasting 1000 steps. We report the reward, which corresponds to the number of steps spent near the final goal of the task automaton, and the success rate ($S.R.$), which is the proportion of policy rollouts that completely satisfied the task specification for all 1000 steps. Because the robustness of rollouts for the final ``Loop'' task cannot be meaningfully evaluated over a finite trajectory \cite{hahn2018omegaregularobjectivesmodelfreereinforcement}, we instead compute the success rate by examining the proportion of rollouts that are never unsafe and successfully complete at least 1 full loop within the episode.

We find that in these environments, stably reaching subgoals with any non-goal-conditioned method is unreliable, especially when rewards are delayed based on the task structure. As a result, \ac{CRM-RS} fails for almost all environments and task types. The \ac{LOF} is able to more easily scale to handle multiple goals due to its hierarchical structure and achieves high robustness for most rollouts. However, it also doesn't typically obtain high rewards due to only learning to apply options for reaching subgoals, as opposed to remaining near the final goal once the task is satisfied. Furthermore, the tasks with safety constraints clearly demonstrate that rollout termination in \ac{CRM-RS} and the reward shaping used by \ac{LOF} is insufficient for reliably preventing the agent from entering obstacle regions. Our method obtained significantly more reward than the baselines, while remaining consistently safe, in all five tasks and environments.

\vspace{-0.1cm}
\subsection{Real-World Experiments}

\begin{wrapfigure}{r}{0.5\linewidth}
    \centering


    \includegraphics[width=0.15\textwidth]{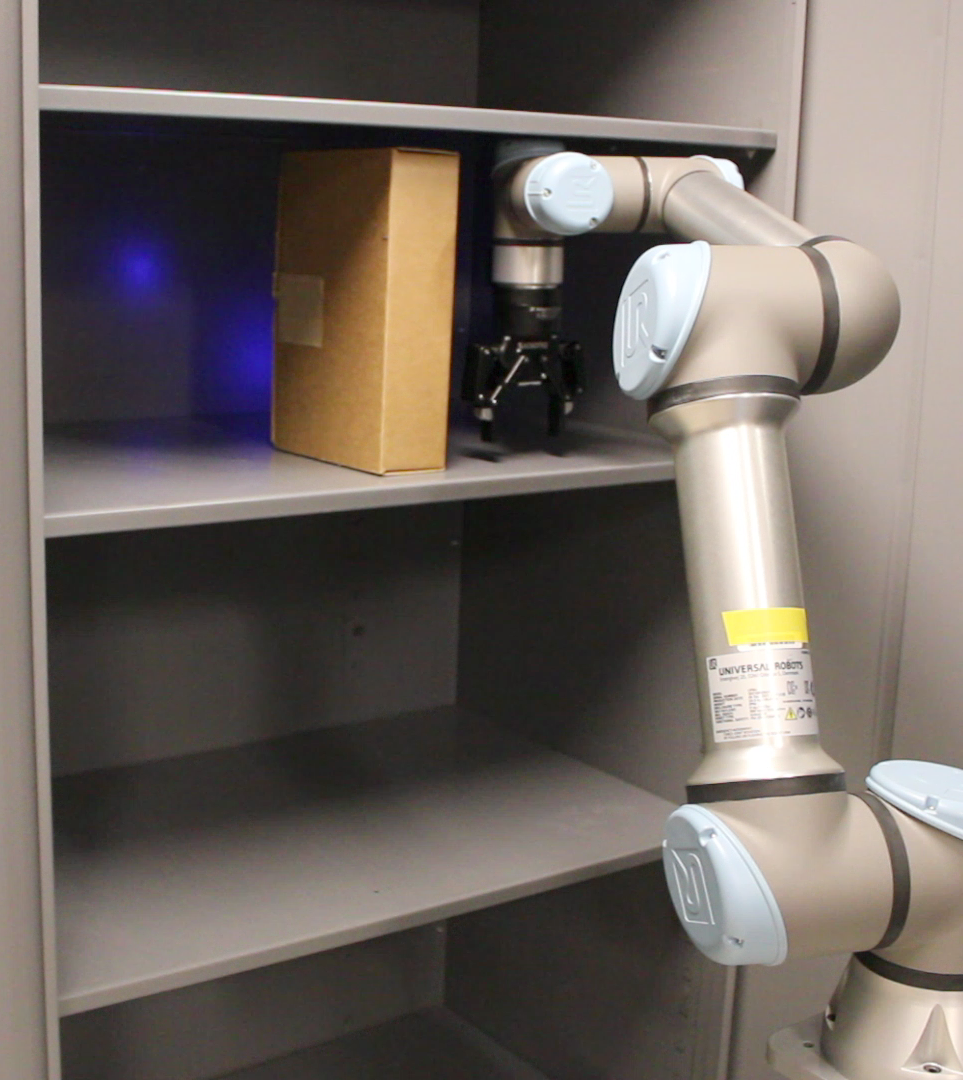}
    \includegraphics[width=0.15\textwidth]{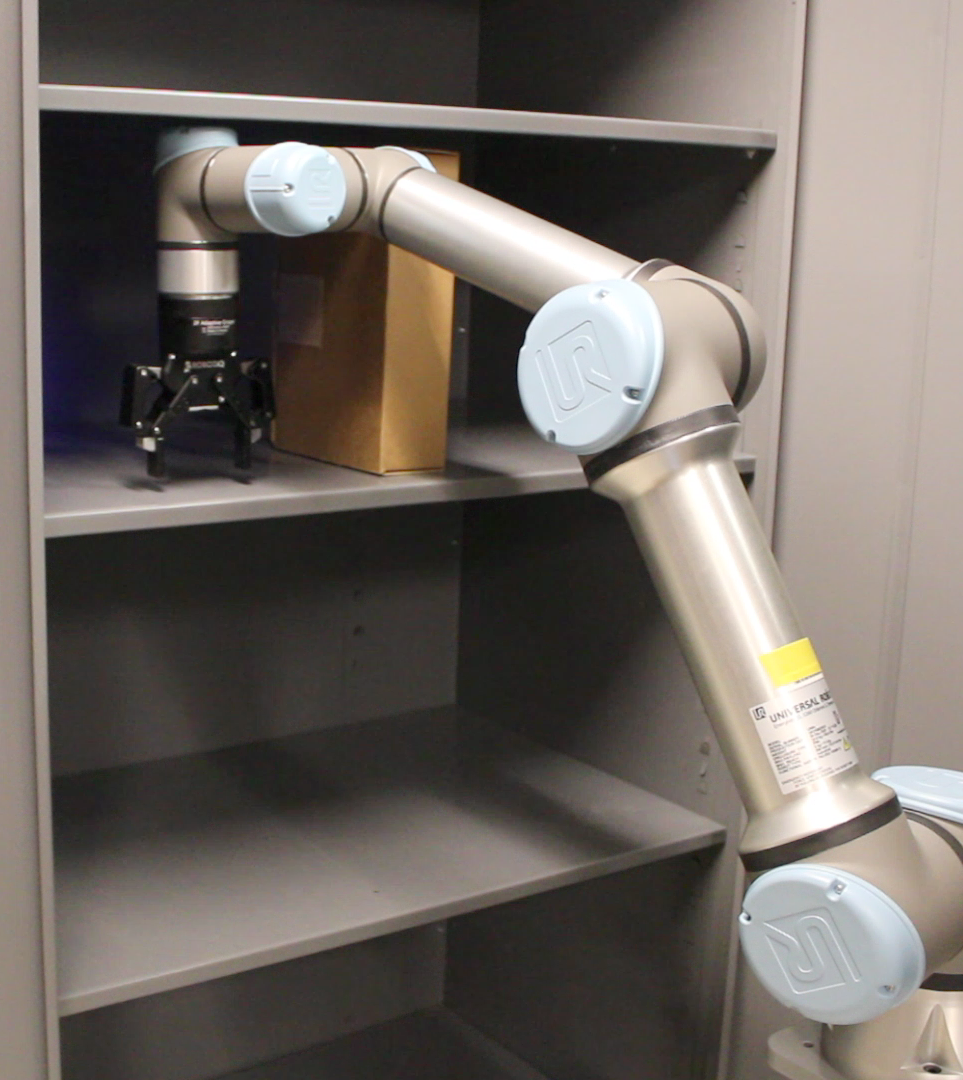}
    \includegraphics[width=0.15\textwidth]{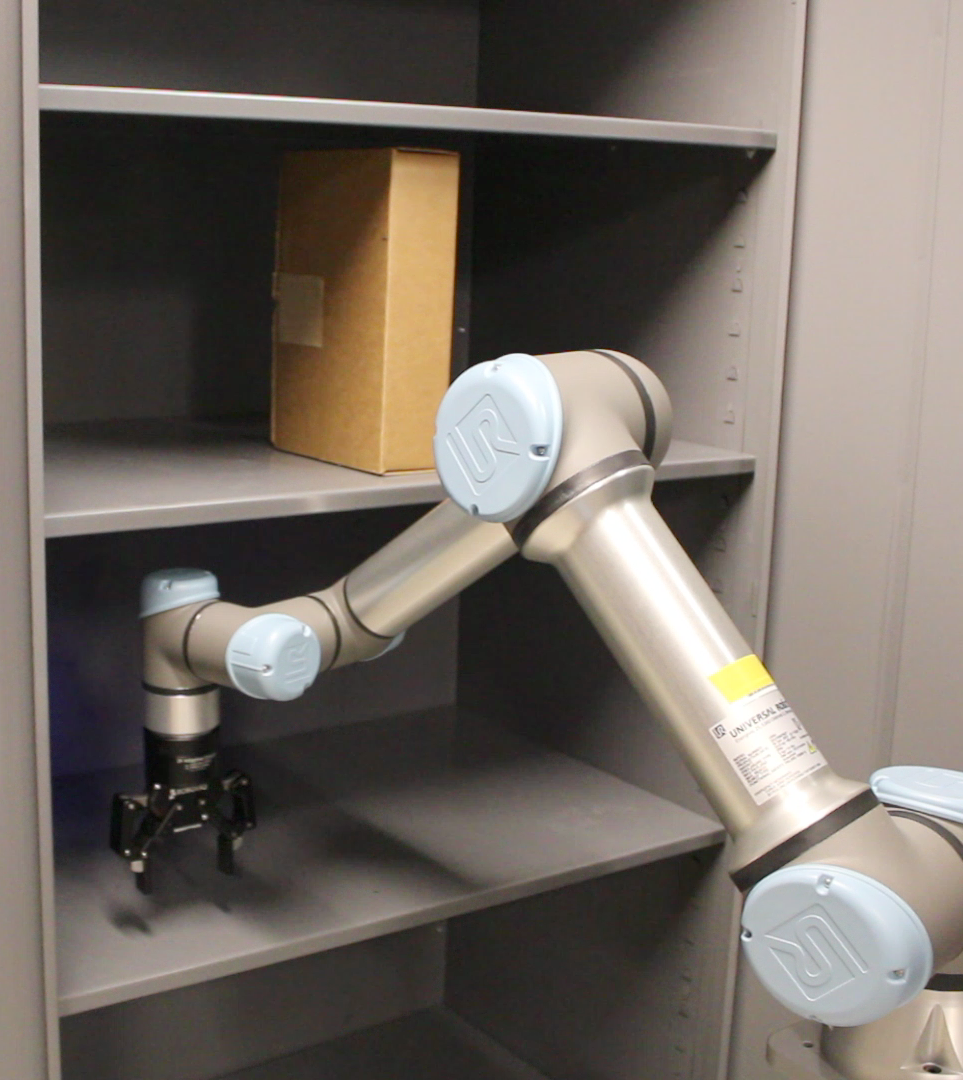}

    \caption{\ac{ACQL} policies trained with safety constraints based a cabinet's geometry can be successfully deployed for a UR5e manipulator operating in the real cabinet environment.}
    \label{fig:real-demo}
\end{wrapfigure}

As a first step in determining our algorithm's applicability to controlling robots in the real world for \ac{LTL} specified tasks, we trained a policy using \ac{ACQL} for controlling a 6-DOF manipulator in a storage cabinet workspace. We trained this policy in the simulated environment pictured in Figure~\ref{fig:environments:ur5e_shelf} for a complex navigation task involving 3 subgoals and 2 obstacles: $\lozenge(p_1 \wedge \circ(\lozenge(p_2 \wedge \circ(\lozenge(p_3))))) \wedge \square(\neg(\text{in\_wall} \vee \text{in\_table}))$. The policy was trained over 6 actions corresponding to translating the end-effector in the 6 cardinal directions. The geometry of the simulated environment perfectly aligned with the real cabinet workspace, so that policy actions based on the state of the simulation could be feasibly executed on the real robot. The resulting policy achieved a mean reward of $908.4$ and a $100\%$ success rate across 16 simulated rollouts, and we successfully deployed this same policy to the real UR5e robot arm and storage cabinet environment visualized in Figure~\ref{fig:real-demo}. To support reproducibility, all relevant details for the task definition and environment setup are provided in our Supplementary Material. In future work, we intend to support a wider sim-to-real gap and train policies for \ac{LTL} objectives in partially observable environments. Nonetheless, these initial results demonstrate that our algorithm can effectively leverage feedback from \ac{STL} tasks in simulation to produce performant and safe policies for real robotic systems.


\subsection{Ablative Analysis}
\label{section:ablations}


\begin{figure}
  \centering
  \begin{minipage}{0.46\textwidth}
    \centering
    \includegraphics[width=\linewidth]{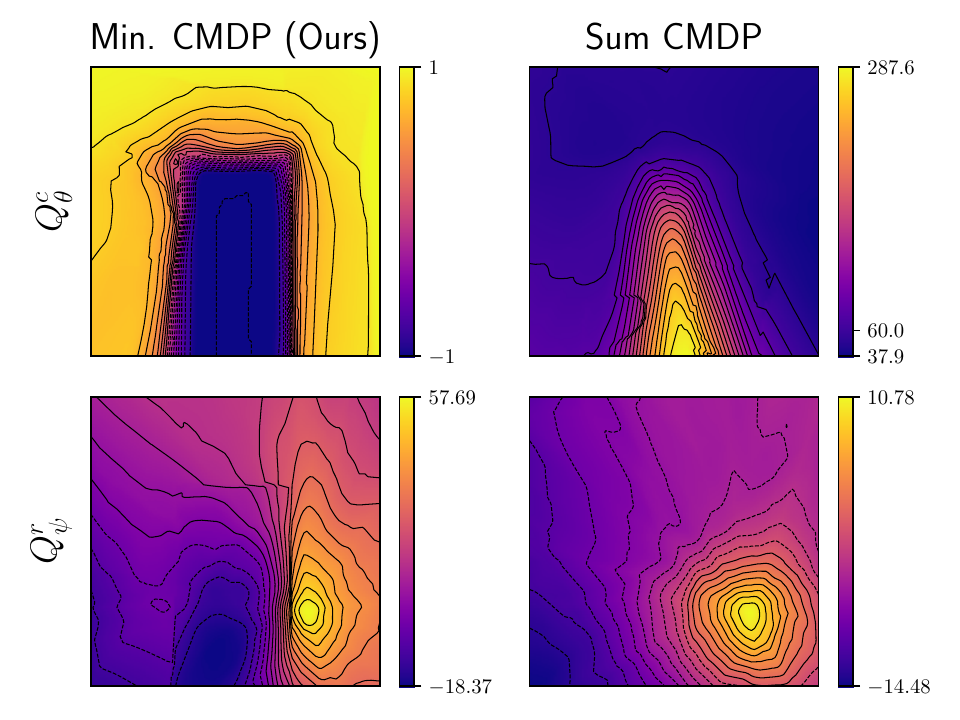}
    \caption{Contour plots for both $Q^c_\theta$ and $Q^r_\psi$ trained with \ac{ACQL} with our \ac{CMDP} formulation in \eqref{eqn:min-cost-cmdp} and the standard formulation in \eqref{eqn:normal-cmdp-objective}.}
    \label{fig:ablation-comparison}
  \end{minipage}
  \hfill
  \begin{minipage}{0.51\textwidth}
    \centering
    \scriptsize
    \captionof{table}{Results for our algorithm when ablating \ac{HER} and substituting our minimum-safety \ac{CMDP} formulation with the standard \ac{CMDP} formulation. We report mean and standard deviation of the reward and success rate collected over all environments, tasks (except the loop task), and seeds from the policy at the end of training in 16 evaluation episodes lasting 1000 steps. For the ablation affecting safety constraints, we only report mean performance in tasks that include safety constraints.}
    \begin{tabular}{llcc}\toprule
        \multicolumn{2}{l}{} & {Reward $\uparrow$} & $S.R.\%$ $\uparrow$ \\
        \midrule
        \multirow{2}{*}[-0.1cm]{\tiny All Tasks} &
        {\tiny ACQL}       &
        \boldsymbol{$682.5 \pm 232.1$} & \boldsymbol{$91.5 \pm 20.3$}                           \\ 
        \cmidrule{2-4}
        & {\tiny No HER}             &
        $95.7 \pm 220.1$              & 
        $12.9 \pm 29.2$                \\ 
        \cmidrule{1-4}
        \multirow{2}{*}[-0.1cm]{\tiny Safety Tasks} &
        {\tiny ACQL} &
        \boldsymbol{$545.8 \pm 365.5$} & \boldsymbol{$75.4 \pm 39.1$} \\
        \cmidrule{2-4}
        & {\tiny Sum \ac{CMDP}} &
        $8.0 \pm 24.9$ & $4.6 \pm 11.0$ \\
        \bottomrule
    \end{tabular}
    \label{tab:ablations}
  \end{minipage}
  \vspace{-12pt}
\end{figure}

We conducted an additional ablative analysis to justify two key components of our algorithm. The results are reported in Table~\ref{tab:ablations}. In the first ablation, we removed the use of \ac{HER} from \ac{ACQL}, requiring the agent to learn exclusively from the sparse reward given upon reaching an accepting automaton state. This ablation isolates the effect of our subgoal-including product \ac{CMDP} formulation that enables applying \ac{GCRL} methods in the context of \ac{LTL} tasks, and our results show that removing it leads to significantly degraded performance.

Our second ablation demonstrates the benefit of our minimum safety constraint \eqref{eqn:min-cost-cmdp} over the discounted sum-of-cost framework \eqref{eqn:normal-cmdp-objective} for accurately learning the state-action safety function. We modified our \ac{CMDP} to provide positive costs when there was a safety violation, trained $Q^c_\theta$ to predict discounted sums of costs, and constrained $\pi$ to choose actions below an upper limit on total cost $\mathcal{L}$. This limit was chosen based on the best performing value $\mathcal{L} \in \{ 0, 10, 40, 60\}$. We report the difference in performance over only the tasks that involved safety constraints and observe that our \ac{CMDP} formulation is critical to the performance of our algorithm. We also show a visualization of the policy differences that result from this ablation in Figure~\ref{fig:ablation-comparison}. We observe that in our formulation (left), \ac{ACQL} can more effectively learn the safety constraint, which substantially improves the quality of $Q^r_\psi$ (bottom). Note that the differing scale for $Q^c_\theta$ in our formulation (top left) results in a yellow safe region, while the standard formulation depicts low-cost, safe regions in blue.


\vspace{-0.2cm}
\section{Conclusion}
\paragraph{Limitations}
\label{section:limitations}
Although our method outperforms baseline methods in training policies to solve several fundamental types of \ac{LTL} tasks, there remain a few limitations that we aim to address in future work. First, like many approaches leveraging product \acp{MDP} \cite{toroicarteRewardMachines2022, quint2021formal, araki2021logicaloptionsframework, tasse2022skill}, our method is restricted to logical tasks representable by \acp{DBA}. To overcome this, we plan to investigate more expressive types of automata for \ac{RL}, such as Good-for-MDPs Non-Deterministic B{\"u}chi Automata \cite{hahn2020good}. Second, our current approach approximates the state-action value function without fully capturing task requirements beyond the current set of subgoals. We aim to explore more sophisticated goal-conditioned \ac{RL} methods that can effectively condition a policy on future subgoals provided by our product \ac{CMDP}. Finally, while our real-world deployment demonstrates the feasibility of applying our algorithm to robot control outside simulation, there are open challenges related to partial observability and sim-to-real mismatch in real-world learning with \ac{LTL} objectives that we aim to address in future work.

We present a novel \ac{RL} algorithm to tackle complex and high-dimensional \ac{LTL}-specified tasks, enabled by a novel augmented product \ac{CMDP} formulation that provides feedback for enforcing arbitrary safety constraints and subgoals for efficiently learning to achieve liveness constraints.
By significantly outperforming comparable baselines for learning from \ac{LTL} specifications, our approach demonstrates a promising trajectory toward learning in increasingly complex and realistic environments, without sacrificing generality in supporting the full expressiveness of \ac{LTL} and other formal languages.


\section*{Acknowledgments}

This material is based upon work supported by the Air Force Office of Scientific Research under award number FA9550-24-1-0233. Any opinions, findings, and conclusions or recommendations expressed in this material are those of the author(s) and do not necessarily reflect the views of the United States Air Force.

\bibliographystyle{unsrt}
\bibliography{references}

@inproceedings{schaul2015uvfa,
  title = 	 {Universal Value Function Approximators},
  author = 	 {Schaul, Tom and Horgan, Daniel and Gregor, Karol and Silver, David},
  booktitle = 	 {Proceedings of the 32nd International Conference on Machine Learning},
  pages = 	 {1312--1320},
  year = 	 {2015},
  editor = 	 {Bach, Francis and Blei, David},
  volume = 	 {37},
  series = 	 {Proceedings of Machine Learning Research},
  address = 	 {Lille, France},
  month = 	 {07--09 Jul},
  publisher =    {PMLR},
  url = 	 {https://proceedings.mlr.press/v37/schaul15.html},
}

@inproceedings{andrychowicz2017her,
 author = {Andrychowicz, Marcin and Wolski, Filip and Ray, Alex and Schneider, Jonas and Fong, Rachel and Welinder, Peter and McGrew, Bob and Tobin, Josh and Pieter Abbeel, OpenAI and Zaremba, Wojciech},
 booktitle = {Advances in Neural Information Processing Systems},
 editor = {I. Guyon and U. Von Luxburg and S. Bengio and H. Wallach and R. Fergus and S. Vishwanathan and R. Garnett},
 pages = {},
 publisher = {Curran Associates, Inc.},
 title = {Hindsight Experience Replay},
 url = {https://proceedings.neurips.cc/paper_files/paper/2017/file/453fadbd8a1a3af50a9df4df899537b5-Paper.pdf},
 volume = {30},
 year = {2017}
}

@book{altman1999constrained,
  title = {Constrained Markov Decision Processes},
  author = {Altman, E.},
  year = {1999},
  series = {Stochastic Modeling Series},
  publisher = {Taylor \& Francis},
  isbn = {978-0-8493-0382-1},
  lccn = {99210415}
}

@InProceedings{achiam2017cpo,
  title = 	 {Constrained Policy Optimization},
  author =       {Joshua Achiam and David Held and Aviv Tamar and Pieter Abbeel},
  booktitle = 	 {Proceedings of the 34th International Conference on Machine Learning},
  pages = 	 {22--31},
  year = 	 {2017},
  editor = 	 {Precup, Doina and Teh, Yee Whye},
  volume = 	 {70},
  series = 	 {Proceedings of Machine Learning Research},
  month = 	 {06--11 Aug},
  publisher =    {PMLR},
  url = 	 {https://proceedings.mlr.press/v70/achiam17a.html},
}

@article{liu2020ipo,
    title={IPO: Interior-Point Policy Optimization under Constraints},
    volume={34},
    url={https://ojs.aaai.org/index.php/AAAI/article/view/5932},
    DOI={10.1609/aaai.v34i04.5932},
    number={04},
    journal={Proceedings of the AAAI Conference on Artificial Intelligence},
    author={Liu, Yongshuai and Ding, Jiaxin and Liu, Xin},
    year={2020},
    month={April},
    pages={4940--4947}
}

@article{jothimurugan2021compositional,
  title={Compositional Reinforcement Learning from Logical Specifications},
  author={Jothimurugan, Kishor and Bansal, Suguman and Bastani, Osbert and Alur, Rajeev},
  journal={Advances in Neural Information Processing Systems},
  volume={34},
  pages={10026--10039},
  year={2021}
}

@inproceedings{li2017reinforcement,
  author={Li, Xiao and Vasile, Cristian-Ioan and Belta, Calin},
  booktitle={2017 IEEE/RSJ International Conference on Intelligent Robots and Systems (IROS)}, 
  title={Reinforcement Learning with Temporal Logic Rewards}, 
  year={2017},
  volume={},
  number={},
  pages={3834--3839},
  doi={10.1109/IROS.2017.8206234}
}

@inproceedings{balakrishnan2019structured,
  author={Balakrishnan, Anand and Deshmukh, Jyotirmoy V.},
  booktitle={2019 IEEE/RSJ International Conference on Intelligent Robots and Systems (IROS)}, 
  title={Structured Reward Shaping using Signal Temporal Logic specifications}, 
  year={2019},
  volume={},
  number={},
  pages={3481--3486},
  doi={10.1109/IROS40897.2019.8968254}
}

@inproceedings{li2018policysearch,
  author={Li, Xiao and Ma, Yao and Belta, Calin},
  booktitle={2018 Annual American Control Conference (ACC)}, 
  title={A Policy Search Method For Temporal Logic Specified Reinforcement Learning Tasks}, 
  year={2018},
  volume={},
  number={},
  pages={240--245},
  doi={10.23919/ACC.2018.8431181}
}

@inproceedings{xiong2024dscrl,
  author={Xiong, Zikang and Lawson, Daniel and Eappen, Joe and Qureshi, Ahmed H. and Jagannathan, Suresh},
  booktitle={2024 IEEE International Conference on Robotics and Automation (ICRA)}, 
  title={Co-learning Planning and Control Policies Constrained by Differentiable Logic Specifications}, 
  year={2024},
  volume={},
  number={},
  pages={14272--14278},
  doi={10.1109/ICRA57147.2024.10610942}
}

@InProceedings{toroicarte2018RewardMachines,
  title = 	 {Using Reward Machines for High-Level Task Specification and Decomposition in Reinforcement Learning},
  author =       {Icarte, Rodrigo Toro and Klassen, Toryn and Valenzano, Richard and McIlraith, Sheila},
  booktitle = 	 {Proceedings of the 35th International Conference on Machine Learning},
  pages = 	 {2107--2116},
  year = 	 {2018},
  editor = 	 {Dy, Jennifer and Krause, Andreas},
  volume = 	 {80},
  series = 	 {Proceedings of Machine Learning Research},
  month = 	 {10--15 Jul},
  publisher =    {PMLR},
  url = 	 {https://proceedings.mlr.press/v80/icarte18a.html},
}

@inproceedings{camacho2019ltl,
  title     = {LTL and Beyond: Formal Languages for Reward Function Specification in Reinforcement Learning},
  author    = {Camacho, Alberto and Toro Icarte, Rodrigo and Klassen, Toryn Q. and Valenzano, Richard and McIlraith, Sheila A.},
  booktitle = {Proceedings of the Twenty-Eighth International Joint Conference on
               Artificial Intelligence, {IJCAI-19}},
  publisher = {International Joint Conferences on Artificial Intelligence Organization},
  pages     = {6065--6073},
  year      = {2019},
  month     = {July},
  doi       = {10.24963/ijcai.2019/840},
  url       = {https://doi.org/10.24963/ijcai.2019/840},
}

@article{toroicarteRewardMachines2022,
   title={Reward Machines: Exploiting Reward Function Structure in Reinforcement Learning},
   volume={73},
   ISSN={1076-9757},
   url={http://dx.doi.org/10.1613/jair.1.12440},
   DOI={10.1613/jair.1.12440},
   journal={Journal of Artificial Intelligence Research},
   publisher={AI Access Foundation},
   author={Toro Icarte, Rodrigo and Klassen, Toryn Q. and Valenzano, Richard and McIlraith, Sheila A.},
   year={2022},
   month=jan,
   pages={173-–208}
}

@inproceedings{araki2021logicaloptionsframework,
  title = 	 {The Logical Options Framework},
  author =       {Araki, Brandon and Li, Xiao and Vodrahalli, Kiran and Decastro, Jonathan and Fry, Micah and Rus, Daniela},
  booktitle = 	 {Proceedings of the 38th International Conference on Machine Learning},
  pages = 	 {307--317},
  year = 	 {2021},
  editor = 	 {Meila, Marina and Zhang, Tong},
  volume = 	 {139},
  series = 	 {Proceedings of Machine Learning Research},
  month = 	 {18--24 Jul},
  publisher =    {PMLR},
  url = 	 {https://proceedings.mlr.press/v139/araki21a.html},
}

@article{li2019formal,
author = {Xiao Li  and Zachary Serlin  and Guang Yang  and Calin Belta },
title = {A formal methods approach to interpretable reinforcement learning for robotic planning},
journal = {Science Robotics},
volume = {4},
number = {37},
pages = {eaay6276},
year = {2019},
doi = {10.1126/scirobotics.aay6276},
URL = {https://www.science.org/doi/abs/10.1126/scirobotics.aay6276},
eprint = {https://www.science.org/doi/pdf/10.1126/scirobotics.aay6276},
}

@inproceedings{tasse2020booleantaskalgebrareinforcement,
    author = {Nangue Tasse, Geraud and James, Steven and Rosman, Benjamin},
    booktitle = {Advances in Neural Information Processing Systems},
    editor = {H. Larochelle and M. Ranzato and R. Hadsell and M.F. Balcan and H. Lin},
    pages = {9497--9507},
    publisher = {Curran Associates, Inc.},
    title = {A Boolean Task Algebra for Reinforcement Learning},
    url = {https://proceedings.neurips.cc/paper_files/paper/2020/file/6ba3af5d7b2790e73f0de32e5c8c1798-Paper.pdf},
    volume = {33},
    year = {2020}
}

@inproceedings{tasse2022skill,
  title={Skill Machines: Temporal Logic Composition in Reinforcement Learning},
  author={Nangue Tasse, Geraud and James, Steven and Rosman, Benjamin},
  booktitle={Deep Reinforcement Learning Workshop @ NeurIPS},
  year={2022}
}

@misc{li2018automata,
      title={Automata-Guided Hierarchical Reinforcement Learning for Skill Composition}, 
      author={Xiao Li and Yao Ma and Calin Belta},
      year={2018},
      eprint={1711.00129},
      archivePrefix={arXiv},
      primaryClass={cs.AI},
      url={https://arxiv.org/abs/1711.00129}, 
}

@inproceedings{hasanbeig2020deep,
  author="Hasanbeig, Mohammadhosein
  and Kroening, Daniel
  and Abate, Alessandro",
  editor="Bertrand, Nathalie
  and Jansen, Nils",
  title="Deep Reinforcement Learning with Temporal Logics",
  booktitle="Formal Modeling and Analysis of Timed Systems",
  year="2020",
  publisher="Springer International Publishing",
  address="Cham",
  pages="1--22",
  isbn="978-3-030-57628-8"
}

@inproceedings{hasanbeig2020cautious,
author = {Hasanbeig, Mohammadhosein and Abate, Alessandro and Kroening, Daniel},
title = {Cautious Reinforcement Learning with Logical Constraints},
year = {2020},
isbn = {9781450375184},
publisher = {International Foundation for Autonomous Agents and Multiagent Systems},
address = {Richland, SC},
booktitle = {Proceedings of the 19th International Conference on Autonomous Agents and MultiAgent Systems},
pages = {483-–491},
numpages = {9},
location = {Auckland, New Zealand},
series = {AAMAS '20}
}

@misc{hasanbeig2019logicallyconstrainedreinforcementlearning,
      title={Logically-Constrained Reinforcement Learning}, 
      author={Mohammadhosein Hasanbeig and Alessandro Abate and Daniel Kroening},
      year={2019},
      eprint={1801.08099},
      archivePrefix={arXiv},
      primaryClass={cs.LG},
      url={https://arxiv.org/abs/1801.08099}, 
}

@misc{quint2021formal,
  title={Formal Language Constrained Markov Decision Processes},
  author={Eleanor Quint and Dong Xu and Samuel W Flint and Stephen D Scott and Matthew Dwyer},
  year={2021},
  url={https://openreview.net/forum?id=NTP9OdaT6nm}
}

@article{alshiekh2018shielding,
title={Safe Reinforcement Learning via Shielding},
volume={32},
url={https://ojs.aaai.org/index.php/AAAI/article/view/11797},
DOI={10.1609/aaai.v32i1.11797},
number={1}, journal={Proceedings of the AAAI Conference on Artificial Intelligence}, author={Alshiekh, Mohammed and Bloem, Roderick and Ehlers, Rüdiger and Könighofer, Bettina and Niekum, Scott and Topcu, Ufuk}, year={2018}, month={April}
}

@misc{kalweit2020deepconstrainedqlearning,
      title={Deep Constrained {Q}-learning}, 
      author={Gabriel Kalweit and Maria Huegle and Moritz Werling and Joschka Boedecker},
      year={2020},
      eprint={2003.09398},
      archivePrefix={arXiv},
      primaryClass={cs.LG},
      url={https://arxiv.org/abs/2003.09398}, 
}

@inproceedings{stooke2020responsivesafetyreinforcementlearning,
    author = {Stooke, Adam and Achiam, Joshua and Abbeel, Pieter},
    title = {Responsive safety in reinforcement learning by PID lagrangian methods},
    year = {2020},
    publisher = {JMLR.org},
    booktitle = {Proceedings of the 37th International Conference on Machine Learning},
    articleno = {847},
    numpages = {11},
    series = {ICML'20}
}

@InProceedings{ha2021saclagrangian,
  title = 	 {Learning to Walk in the Real World with Minimal Human Effort},
  author =       {Ha, Sehoon and Xu, Peng and Tan, Zhenyu and Levine, Sergey and Tan, Jie},
  booktitle = 	 {Proceedings of the 2020 Conference on Robot Learning},
  pages = 	 {1110--1120},
  year = 	 {2021},
  editor = 	 {Kober, Jens and Ramos, Fabio and Tomlin, Claire},
  volume = 	 {155},
  series = 	 {Proceedings of Machine Learning Research},
  month = 	 {16--18 Nov},
  publisher =    {PMLR},
  url = 	 {https://proceedings.mlr.press/v155/ha21c.html}
}

@inproceedings{pnueli1977ltl,
  author={Amir Pnueli},
  booktitle={18th Annual Symposium on Foundations of Computer Science (SFCS 1977)}, 
  title={The Temporal Logic of Programs}, 
  year={1977},
  volume={},
  number={},
  pages={46--57},
  doi={10.1109/SFCS.1977.32}
}

@article{alpern1987Recognizing,
	author = {Bowen Alpern and Fred B. Schneider},
	title = {Recognizing safety and liveness},
	volume = {2},
	issn = {1432-0452},
	url = {https://doi.org/10.1007/BF01782772},
	doi = {10.1007/BF01782772},
	number = {3},
	journal = {Distributed Computing},
	month = sep,
	year = {1987},
	pages = {117--126},
}

@book{baier2008principles,
  title = {Principles of Model Checking},
  author = {Baier, C. and Katoen, J.P. and Larsen, K.G.},
  year = {2008},
  publisher = {MIT Press},
  isbn = {978-0-262-02649-9},
  lccn = {2007037603}
}

@book{borkar2008stochasticapproximation,
  title = { Stochastic Approximation: A Dynamical Systems Viewpoint},
  author = {Borkar, Vivek S.},
  year = {2008},
  publisher = {Hindustan Book Agency Gurgaon},
  isbn = {978-93-86279-38-5},
}

@inproceedings{hahn2020good,
  title = {Good-for-Mdps Automata for Probabilistic Analysis and Reinforcement Learning},
  booktitle = {Tools and Algorithms for the Construction and Analysis of Systems},
  author = {Hahn, Ernst Moritz and Perez, Mateo and Schewe, Sven and Somenzi, Fabio and Trivedi, Ashutosh and Wojtczak, Dominik},
  editor = {Biere, Armin and Parker, David},
  year = {2020},
  pages = {306--323},
  publisher = {Springer International Publishing},
  address = {Cham},
  isbn = {978-3-030-45190-5}
}

@misc{schulman2017proximalpolicyoptimizationalgorithms,
      title={Proximal Policy Optimization Algorithms}, 
      author={John Schulman and Filip Wolski and Prafulla Dhariwal and Alec Radford and Oleg Klimov},
      year={2017},
      eprint={1707.06347},
      archivePrefix={arXiv},
      primaryClass={cs.LG},
      url={https://arxiv.org/abs/1707.06347}, 
}

@INPROCEEDINGS{dawson2022robust_stl_planning,
  author={Dawson, Charles and Fan, Chuchu},
  booktitle={2022 IEEE/RSJ International Conference on Intelligent Robots and Systems (IROS)}, 
  title={Robust Counterexample-guided Optimization for Planning from Differentiable Temporal Logic}, 
  year={2022},
  volume={},
  number={},
  pages={7205--7212},
  doi={10.1109/IROS47612.2022.9981382}}

@inproceedings{duret22cav,
  author = {Alexandre~Duret-Lutz and Etienne Renault and Maximilien Colange and Florian Renkin and Alexandre Gbaguidi~Aisse and Philipp Schlehuber-Caissier and Thomas Medioni and Antoine Martin and J{\'e}r{\^o}me Dubois and Cl{\'e}ment Gillard and Henrich Lauko},
  title = {From {S}pot 2.0 to {S}pot 2.10: What's New?},
  booktitle = {Proceedings of the 34th International Conference on Computer Aided Verification (CAV'22)},
  year = 2022,
  volume = {13372},
  series = {Lecture Notes in Computer Science},
  pages = {174--187},
  month = aug,
  publisher = {Springer},
  slides = {adl/duret.22.cav.slides.pdf},
  doi = {10.1007/978-3-031-13188-2_9}
}

@inproceedings{fisac2019bridging,
  author={Fisac, Jaime F. and Lugovoy, Neil F. and Rubies-Royo, Vicenç and Ghosh, Shromona and Tomlin, Claire J.},
  booktitle={2019 International Conference on Robotics and Automation (ICRA)}, 
  title={Bridging Hamilton-Jacobi Safety Analysis and Reinforcement Learning}, 
  year={2019},
  volume={},
  number={},
  pages={8550--8556},
  doi={10.1109/ICRA.2019.8794107}
}

@inproceedings{hsu2021safety,
    AUTHOR    = {Kai-Chieh Hsu and Vicenç Rubies-Royo and Claire J. Tomlin and Jaime F. Fisac},
    TITLE     = {Safety and Liveness Guarantees through Reach-Avoid Reinforcement Learning},
    BOOKTITLE = {Proceedings of Robotics: Science and Systems},
    YEAR      = {2021},
    ADDRESS   = {Held Virtually},
    MONTH     = {July},
    DOI       = {10.15607/RSS.2021.XVII.077}
}

@software{brax2021github,
  author = {C. Daniel Freeman and Erik Frey and Anton Raichuk and Sertan Girgin and Igor Mordatch and Olivier Bachem},
  title = {Brax - A Differentiable Physics Engine for Large Scale Rigid Body Simulation},
  url = {http://github.com/google/brax},
  version = {0.11.0},
  year = {2021},
}

@article{bortkiewicz2024accelerating,
  title   = {Accelerating Goal-Conditioned {RL} Algorithms and Research},
  author  = {Michał Bortkiewicz and Władek Pałucki and Vivek Myers and Tadeusz Dziarmaga and Tomasz Arczewski and Łukasz Kuciński and Benjamin Eysenbach},
  year    = {2024},
  journal = {arXiv preprint arXiv: 2408.11052}
}

@article{haarnoja2018soft,
  title={Soft actor-critic: Off-policy maximum entropy deep reinforcement learning with a stochastic actor},
  author={Haarnoja, Tuomas and Zhou, Aurick and Abbeel, Pieter and Levine, Sergey},
  journal={International Conference on Machine Learning (ICML)},
  year={2018}
}

@article{alpern1985liveness,
    title = {Defining liveness},
    author = {Bowen Alpern and Fred Schneider},
    journal = {Information Processing Letters},
    volume = {21},
    number = {4},
    pages = {181--185},
    year = {1985},
    issn = {0020-0190},
    doi = {https://doi.org/10.1016/0020-0190(85)90056-0},
}

@inproceedings{manna1990hierarchy,
    author = {Manna, Zohar and Pnueli, Amir},
    title = {A hierarchy of temporal properties (invited paper, 1989)},
    year = {1990},
    isbn = {089791404X},
    publisher = {Association for Computing Machinery},
    address = {New York, NY, USA},
    url = {https://doi.org/10.1145/93385.93442},
    doi = {10.1145/93385.93442},
    booktitle = {Proceedings of the Ninth Annual ACM Symposium on Principles of Distributed Computing},
    pages = {377-–410},
    numpages = {34},
    location = {Quebec City, Quebec, Canada},
    series = {PODC '90}
}

@INPROCEEDINGS{aksaray2016qlearningrobustsatisfactionsignal,
  author={Aksaray, Derya and Jones, Austin and Kong, Zhaodan and Schwager, Mac and Belta, Calin},
  booktitle={2016 IEEE 55th Conference on Decision and Control (CDC)}, 
  title={{Q}-Learning for robust satisfaction of signal temporal logic specifications}, 
  year={2016},
  volume={},
  number={},
  pages={6565--6570},
  doi={10.1109/CDC.2016.7799279}}

@inproceedings{qiu2023instructing,
 author = {Qiu, Wenjie and Mao, Wensen and Zhu, He},
 booktitle = {Advances in Neural Information Processing Systems},
 editor = {A. Oh and T. Naumann and A. Globerson and K. Saenko and M. Hardt and S. Levine},
 pages = {39147--39175},
 publisher = {Curran Associates, Inc.},
 title = {Instructing Goal-Conditioned Reinforcement Learning Agents with Temporal Logic Objectives},
 url = {https://proceedings.neurips.cc/paper_files/paper/2023/file/7b35a69f434b5eb07ed1b1ef16ace52c-Paper-Conference.pdf},
 volume = {36},
 year = {2023}
}

@inproceedings{kaelbling1993learning,
  title={Learning to achieve goals},
  author={Kaelbling, Leslie Pack},
  booktitle={International Joint Conference on Artificial Intelligence},
  volume={2},
  pages={1094--8},
  year={1993},
  organization={Citeseer}
}

@inproceedings{schaul2015universal,
  title={Universal value function approximators},
  author={Schaul, Tom and Horgan, Daniel and Gregor, Karol and Silver, David},
  booktitle={International Conference on Machine Learning},
  pages={1312--1320},
  year={2015},
  organization={PMLR}
}

@InProceedings{hahn2018omegaregularobjectivesmodelfreereinforcement,
    author="Hahn, Ernst Moritz
    and Perez, Mateo
    and Schewe, Sven
    and Somenzi, Fabio
    and Trivedi, Ashutosh
    and Wojtczak, Dominik",
    editor="Vojnar, Tom{\'a}{\v{s}}
    and Zhang, Lijun",
    title="Omega-Regular Objectives in Model-Free Reinforcement Learning",
    booktitle="Tools and Algorithms for the Construction and Analysis of Systems",
    year="2019",
    publisher="Springer International Publishing",
    address="Cham",
    pages="395--412",
    isbn="978-3-030-17462-0"
}

@InProceedings{yu2022reachabilityconstrainedreinforcementlearning,
  title = 	 {Reachability Constrained Reinforcement Learning},
  author =       {Yu, Dongjie and Ma, Haitong and Li, Shengbo and Chen, Jianyu},
  booktitle = 	 {Proceedings of the 39th International Conference on Machine Learning},
  pages = 	 {25636--25655},
  year = 	 {2022},
  editor = 	 {Chaudhuri, Kamalika and Jegelka, Stefanie and Song, Le and Szepesvari, Csaba and Niu, Gang and Sabato, Sivan},
  volume = 	 {162},
  series = 	 {Proceedings of Machine Learning Research},
  month = 	 {17--23 Jul},
  publisher =    {PMLR},
  url = 	 {https://proceedings.mlr.press/v162/yu22d.html},
}

@misc{chow2017riskconstrainedreinforcementlearningpercentile,
      title={Risk-Constrained Reinforcement Learning with Percentile Risk Criteria}, 
      author={Yinlam Chow and Mohammad Ghavamzadeh and Lucas Janson and Marco Pavone},
      year={2017},
      eprint={1512.01629},
      archivePrefix={arXiv},
      primaryClass={cs.AI},
      url={https://arxiv.org/abs/1512.01629}, 
}

@InProceedings{ma2022jointsynthesissafetycertificate,
  title = 	 {Joint Synthesis of Safety Certificate and Safe Control Policy Using Constrained Reinforcement Learning},
  author =       {Ma, Haitong and Liu, Changliu and Li, Shengbo Eben and Zheng, Sifa and Chen, Jianyu},
  booktitle = 	 {Proceedings of The 4th Annual Learning for Dynamics and Control Conference},
  pages = 	 {97--109},
  year = 	 {2022},
  editor = 	 {Firoozi, Roya and Mehr, Negar and Yel, Esen and Antonova, Rika and Bohg, Jeannette and Schwager, Mac and Kochenderfer, Mykel},
  volume = 	 {168},
  series = 	 {Proceedings of Machine Learning Research},
  month = 	 {23--24 Jun},
  publisher =    {PMLR},
  url = 	 {https://proceedings.mlr.press/v168/ma22a.html},
}

@article{tsitsiklis1994asynchronous,
    author = {Tsitsiklis, John N.},
    title = {Asynchronous Stochastic Approximation and {Q}-Learning},
    year = {1994},
    issue_date = {Sept. 1994},
    publisher = {Kluwer Academic Publishers},
    address = {USA},
    volume = {16},
    number = {3},
    issn = {0885-6125},
    url = {https://doi.org/10.1023/A:1022689125041},
    doi = {10.1023/A:1022689125041},
    journal = {Machine Learning},
    month = sep,
    pages = {185–202},
    numpages = {18}
}

@book{granas2003fixed,
  title={Fixed Point Theory},
  author={Granas, A. and Dugundji, J.},
  isbn={9780387001739},
  lccn={2002042736},
  series={Monographs in Mathematics},
  url={https://books.google.com/books?id=4_iJAoLSq3cC},
  year={2003},
  publisher={Springer}
}

@inproceedings{wang2023hardconstraints,
author = {Wang, Yixuan and Zhan, Simon Sinong and Jiao, Ruochen and Wang, Zhilu and Jin, Wanxin and Yang, Zhuoran and Wang, Zhaoran and Huang, Chao and Zhu, Qi},
title = {Enforcing hard constraints with soft barriers: safe reinforcement learning in unknown stochastic environments},
year = {2023},
publisher = {JMLR.org},
booktitle = {Proceedings of the 40th International Conference on Machine Learning},
articleno = {1522},
numpages = {12},
location = {Honolulu, Hawaii, USA},
series = {ICML'23}
}

@book{bertsekas2015parallel,
  title={Parallel and Distributed Computation: Numerical Methods},
  author={Bertsekas, D. and Tsitsiklis, J.},
  isbn={9781886529151},
  lccn={2014070648},
  url={https://books.google.com/books?id=n_Q5EAAAQBAJ},
  year={2015},
  publisher={Athena Scientific}
}

@article{konighofer2023online,
  title =        {Online shielding for reinforcement learning},
  volume =       {19},
  issn =         {1614-5054},
  url =          {https://doi.org/10.1007/s11334-022-00480-4},
  doi =          {10.1007/s11334-022-00480-4},
  number =       {4},
  journal =      {Innovations in Systems and Software Engineering},
  author =       {Könighofer, Bettina and Rudolf, Julian and Palmisano,
                  Alexander and Tappler, Martin and Bloem, Roderick},
  month =        dec,
  year =         {2023},
  pages =        {379--394},
}

@inproceedings{desai2017combining,
  address =      {Cham},
  title =        {Combining Model Checking and Runtime Verification for
                  Safe Robotics},
  isbn =         {978-3-319-67531-2},
  booktitle =    {Runtime Verification},
  publisher =    {Springer International Publishing},
  author =       {Desai, Ankush and Dreossi, Tommaso and Seshia, Sanjit A.},
  editor =       {Lahiri, Shuvendu and Reger, Giles},
  year =         {2017},
  pages =        {172--189},
}

@inproceedings{jackermeier2025deepltl,
title={Deep{LTL}: Learning to Efficiently Satisfy Complex {LTL} Specifications for Multi-Task {RL}},
author={Mathias Jackermeier and Alessandro Abate},
booktitle={The Thirteenth International Conference on Learning Representations},
year={2025},
url={https://openreview.net/forum?id=9pW2J49flQ}
}

@article{shah2025ltlconstrained,
title={{LTL}-Constrained Policy Optimization with Cycle Experience Replay},
author={Ameesh Shah and Cameron Voloshin and Chenxi Yang and Abhinav Verma and Swarat Chaudhuri and Sanjit A. Seshia},
journal={Transactions on Machine Learning Research},
issn={2835-8856},
year={2025},
url={https://openreview.net/forum?id=gxUp2d4JTw},
note={}
}

@article{xu2024generalization,
	title = {Generalization of temporal logic tasks via future dependent options},
	volume = {113},
	issn = {1573-0565},
	url = {https://doi.org/10.1007/s10994-024-06614-y},
	doi = {10.1007/s10994-024-06614-y},
	number = {10},
	journal = {Machine Learning},
	author = {Xu, Duo and Fekri, Faramarz},
	month = oct,
	year = {2024},
	pages = {7509--7540},
}

@inproceedings{yalcinkaya2023automata,
title={Automata Conditioned Reinforcement Learning with Experience Replay},
author={Beyazit Yalcinkaya and Niklas Lauffer and Marcell Vazquez-Chanlatte and Sanjit Seshia},
booktitle={NeurIPS 2023 Workshop on Goal-Conditioned Reinforcement Learning},
year={2023},
url={https://openreview.net/forum?id=ktyvmR8btd}
}

@InProceedings{vaezipoor2021ltl2action,
  title = 	 {{LTL}2{A}ction: Generalizing {LTL} Instructions for Multi-Task {RL}},
  author =       {Vaezipoor, Pashootan and Li, Andrew C and Icarte, Rodrigo A Toro and Mcilraith, Sheila A.},
  booktitle = 	 {Proceedings of the 38th International Conference on Machine Learning},
  pages = 	 {10497--10508},
  year = 	 {2021},
  editor = 	 {Meila, Marina and Zhang, Tong},
  volume = 	 {139},
  series = 	 {Proceedings of Machine Learning Research},
  month = 	 {18--24 Jul},
  publisher =    {PMLR},
  url = 	 {https://proceedings.mlr.press/v139/vaezipoor21a.html},
}

@InProceedings{yalcinkaya2025automataconditionedrl,
  title = 	 {Provably Correct Automata Embeddings for Optimal Automata-Conditioned Reinforcement Learning},
  author =       {Yalcinkaya, Beyazit and Lauffer, Niklas and Vazquez-Chanlatte, Marcell and Seshia, Sanjit A.},
  booktitle = 	 {Proceedings of the International Conference on Neuro-symbolic Systems},
  pages = 	 {661--675},
  year = 	 {2025},
  editor = 	 {Pappas, George and Ravikumar, Pradeep and Seshia, Sanjit A.},
  volume = 	 {288},
  series = 	 {Proceedings of Machine Learning Research},
  month = 	 {28--30 May},
  publisher =    {PMLR},
  url = 	 {https://proceedings.mlr.press/v288/yalcinkaya25a.html},
}

@inproceedings{yalcinkaya2024compositional,
	title = {Compositional Automata Embeddings for Goal-Conditioned Reinforcement Learning},
	volume = {37},
	url = {https://proceedings.neurips.cc/paper_files/paper/2024/file/858fc542b70d3b39067f7d3b1cd77635-Paper-Conference.pdf},
	booktitle = {Advances in {Neural} {Information} {Processing} {Systems}},
	publisher = {Curran Associates, Inc.},
	author = {Yalcinkaya, Beyazit and Lauffer, Niklas and Vazquez-Chanlatte, Marcell and Seshia, Sanjit A.},
	editor = {Globerson, A. and Mackey, L. and Belgrave, D. and Fan, A. and Paquet, U. and Tomczak, J. and Zhang, C.},
	year = {2024},
	pages = {72933--72963},
}

\begin{acronym}[TDMA]
\acro{ACQL}{Automaton Constrained Q-Learning}
\acro{RL}{Reinforcement Learning}
\acro{TL}{Temporal Logic}
\acro{HJ}{Hamilton-Jacobi}
\acro{GCRL}{Goal-Conditioned Reinforcement Learning}
\acro{RM}{Reward Machine}
\acro{HER}{Hindsight Experience Replay}
\acro{LTL}{Linear-time Temporal Logic}
\acro{STL}{Signal Temporal Logic}
\acro{HRL}{Hierarchical Reinforcement Learning}
\acro{MDP}{Markov Decision Process}
\acro{CMDP}{Constrained Markov Decision Process}
\acro{DBA}{Deterministic B{\"u}chi Automaton}
\acrodefplural{DBA}{Deterministic B{\"u}chi Automata}
\acro{CRM}{Counterfactual Experiences for Reward Machines}
\acro{CRM-RS}{Counterfactual Experiences for Reward Machines with Reward Shaping}
\acro{DQN}{Deep Q-Networks}
\acro{LOF}{Logical Options Framework}
\acro{PPO}{Proximal Policy Optimization}
\acro{DDPG}{Deep Deterministic Policy Gradient}
\acro{GAS}{Globally Asymptotically Stable}
\acrodefplural{MDP}{Markov Decision Processes}
\acrodefplural{CMDP}{Constrained Markov Decision Processes}
\end{acronym}


\newpage

\appendix



\section{Proof of Proposition~\ref{proposition:convergence}}
\label{appendix:proof}

In what follows, we formally prove and restate Proposition~\ref{proposition:convergence}, which shows that \ac{ACQL}, under mild conditions, is guaranteed to return the optimal policy. Our proof is based on the proof of convergence for Q-learning using stochastic approximation theory in \cite{tsitsiklis1994asynchronous}, extended with the theory of stochastic approximation under multiple timescales described in Chapter 6 of \cite{borkar2008stochasticapproximation}. Refer also to \cite{chow2017riskconstrainedreinforcementlearningpercentile, ma2022jointsynthesissafetycertificate, yu2022reachabilityconstrainedreinforcementlearning} for similar analyses. The outline of the section is as follows:


\begin{enumerate}
    \item Express \ac{ACQL} as a stochastic approximation algorithm \cite{tsitsiklis1994asynchronous, borkar2008stochasticapproximation}.
    \item Show that $Q^c$ and $Q^r$ converge to an optimal fixed point for any fixed safety discount factor $\gamma_c \in (0, 1)$.
    \item Restate our Proposition~\ref{proposition:convergence} and prove it by showing that as $\gamma_c \rightarrow 1$, $Q^r$ and $Q^c$ converge to the optimal state-action value function $Q^{r*}$ and its corresponding optimal (undiscounted) state-action safety function $Q^{c*}$.
\end{enumerate}



\subsection{Setup}

\begin{assumption}
    \label{assumption:finite-mdp}
    The augmented \ac{CMDP} $\mathcal{M}^{A} = (\mathcal{S}^{A}, \mathcal{A}, \mathcal{T}^{A}, d_0^{A}, r^{A}, c^{A}, \gamma, \mathcal{L})$ for \ac{ACQL} is defined on a finite state space $\mathcal{S}^A$ and action space $\mathcal{A}$. For every state $s \in \mathcal{S}^A$ and action $a \in \mathcal{A}$, there is an associated bounded deterministic reward $r_{sa} = r^A(s, a)$ and bounded constraint feedback $c_{sa} = c^A(s, a)$ observed if action $a$ is applied at state $s$.
\end{assumption}

Under Assumption~\ref{assumption:finite-mdp}, $Q^c$ and $Q^r$ are vectors in $\mathbb{R}^d$ where $d = | \mathcal{S}^A \times \mathcal{A}|$ is finite. The \ac{ACQL} algorithm can be modeled as a distributed, asynchronous series of noisy updates to components of $Q^c$ and $Q^r$.

\begin{assumption}
    \label{assumption:infinite-updates}
    For each state-action pair $(s, a) \in \mathcal{S}^A \times \mathcal{A}$, there are an infinite number of updates applied to the components $Q^r_{s,a}$ and $Q^c_{s,a}$.
\end{assumption}


The updates to these components are given by
\begin{align}
    \label{eqn:qc-update}
    Q^c_{s,a}(n+1) &= Q^c_{s,a}(n) + a(n) \left [ \left ( (1 - \gamma_c)c_{sa} + \gamma_c \min \{ c_{sa}, \bar{Q}^c_{s',\pi(s')}(n) \} \right ) - Q^c_{s,a}(n) \right ]   \; \text{and} \\
    \label{eqn:qr-update}
    Q^r_{s,a}(n+1) &= Q^r_{s,a}(n) + b(n) \left [ (r_{sa} + \gamma \bar{Q}^r_{s',\pi(s')}(n) ) - Q^r_{s,a}(n) \right ],
\end{align}
where $s'$ is a randomly sampled next state following the state $s$ and action $a$. The elements of $\bar{Q}^c(n)$ and $\bar{Q}^r(n)$ are potentially taken from older iterations $Q^c_{s,a}(\nu_{s,a}(n))$ and $Q^r_{s,a}(\nu_{s,a}(n))$ where $\nu_{s,a}(n)$ is an integer satisfying $0 \leq \nu_{s,a}(n) \leq n$. Recall that the policy $\pi$ here is defined as $\argmax_{a : \bar{Q}^c_{s, a} > \mathcal{L}} \bar{Q}^r_{s, a}$ in terms of $\bar{Q}^c$ and $\bar{Q}^r$. However, we assume that old information is eventually discarded as $n \rightarrow \infty$.
\begin{assumption}
    \label{assumption:eventually-discarding-info}
    For all $(s,a)$, $\lim_{n \rightarrow \infty} \nu_{s,a}(n) = \infty$.
\end{assumption}
Assumption~\ref{assumption:eventually-discarding-info} is necessary to prove the convergence of distributed asynchronous stochastic approximation algorithms using outdated values ($\bar{Q}^c$ and $\bar{Q}^r$) \cite{tsitsiklis1994asynchronous, bertsekas2015parallel}. Using $\bar{Q}^c$ is additionally necessary to define the operator in \eqref{eqn:qc-operator} such that a fixed policy $\pi$ can be used to prove the contraction property in Lemma~\ref{lemma:g-contraction-property}.

We also update $\gamma_c$ infinitely often with
\begin{align}
    \label{eqn:gamma-c-update}
    \gamma_c(n+1) &= \gamma_c(n) + c(n) \left [ \left ( 1 - \gamma_c(n) \right ) - \gamma_c(n) \right ].
\end{align}
and assume that its updates are synchronized with the index $n$ for the updates to the components of $Q^c$ and $Q^r$.
\begin{assumption}
    \label{assumption:robbins-monro-separated-step-sizes}
    The step sizes $a(n)$, $b(n)$, and $c(n)$ for the above updates satisfy
    \begin{align*}
        \sum_{n=0}^{\infty} a(n) = \sum_{n=0}^{\infty} b(n) = \sum_{n=0}^{\infty} c(n) &= \infty \\
        \sum_{n=0}^{\infty} a(n)^2, \sum_{n=0}^{\infty} b(n)^2, \sum_{n=0}^{\infty} c(n)^2 &< \infty,
    \end{align*}
    $b(n) \in o(a(n))$, and $c(n) \in o(b(n))$.
\end{assumption}



Now, let $g : \mathbb{R}^d \times \mathbb{R}^d \times \mathbb{R} \rightarrow \mathbb{R}^d$ and $h : \mathbb{R}^d \times \mathbb{R}^d \times \mathbb{R} \rightarrow \mathbb{R}^d$ be operators defined for each component $(s,a)$ as
\begin{align}
    \label{eqn:qc-operator}
    g_{s,a}(Q^r, Q^c, \gamma_c) &= (1 - \gamma_c)c_{sa} + \gamma_c \expectation_{s'} \left [ \min \{ c_{sa}, Q^c_{s',\pi(s')} \} \right ] \\
    h_{s,a}(Q^r, Q^c, \gamma_c) &= r_{sa} + \gamma \expectation_{s'} \left [ Q^r_{s',\pi(s')} \right ].
\end{align}
Define a third mapping $f(Q^r, Q^c, \gamma_c) = 1 - \gamma_c$. Without loss of generality, we express all the operators as mappings from $\mathbb{R}^d \times \mathbb{R}^d \times \mathbb{R}$ to consider them as a coupled mapping from $\mathbb{R}^{2d+1}$ to $\mathbb{R}^{2d+1}$. Finally, define two martingale difference sequences
\begin{align}
    \label{eqn:mc-martingale-difference-sequence}
    M^c_{s,a}(n+1) &= \gamma_c \min \{ c_{sa}, \bar{Q}^c_{s',\pi(s')}(n) \} - \gamma_c \expectation_{s'} \left [ \min \{ c_{sa}, \bar{Q}^c_{s',\pi(s')}(n) \} \right ]  \; \text{and} \\
    \label{eqn:mr-martingale-difference-sequence}
    M^r_{s,a}(n+1) &= \gamma \bar{Q}^r_{s',\pi(s')}(n) - \gamma \expectation_{s'} \left [ \bar{Q}^r_{s',\pi(s')}(n) \right ].
\end{align}
The updates to $\gamma^c$ are deterministic. Using the above, we can express the three \ac{ACQL} updates \eqref{eqn:qc-update}, \eqref{eqn:qr-update}, and \eqref{eqn:gamma-c-update} as
\begin{align}
    \label{eqn:stochastic-qc-update}
    Q^c_{s,a}(n+1) &= Q^c_{s,a}(n) + a(n) \left [ g_{s,a}(\bar{Q^r}(n), \bar{Q^c}(n), \gamma_c(n)) - Q^c_{s,a}(n) + M^c_{s,a}(n+1) \right ], \\
    \label{eqn:stochastic-qr-update}
    Q^r_{s,a}(n+1) &= Q^r_{s,a}(n) + b(n) \left [ h_{s,a}(\bar{Q^r}(n), \bar{Q^c}(n), \gamma_c(n)) - Q^r_{s,a}(n) + M^r_{s,a}(n+1) \right ], \\
    \label{eqn:stochastic-gamma-c-update}
    \gamma_c(n+1) &= \gamma_c(n) + c(n) \left [ f(Q^r(n), Q^c(n), \gamma_c(n)) - \gamma_c(n) \right ].
\end{align}
Under the above assumptions and formulation, \ac{ACQL} can now be analyzed as a distributed stochastic approximation algorithm under three timescales.

\subsection{Convergence of $Q^r$ and $Q^c$ under a fixed $\gamma_c$}

Since the update to $\gamma_c$ happens much more slowly than the updates to $Q^r$ and $Q^c$---formally, the step size $c(n)$ for $\gamma_c$ shrinks faster than both $b(n)$ and $a(n)$---we can treat $\gamma_c$ as approximately fixed while analyzing the behavior of $Q^r$ and $Q^c$. This allows us to study the convergence of $Q^r$ and $Q^c$ assuming that $\gamma_c$ is a constant value in the interval $(0, 1)$.

\begin{lemma}
    \label{lemma:g-contraction-property}
    The mapping $g^c = g(Q^r, \cdot, \gamma_c) : \mathbb{R}^d \rightarrow \mathbb{R}^d$, for some fixed $Q^r$, some fixed feasible policy $\pi$ (e.g., a policy based on $\bar{Q}^c$ and $\bar{Q}^r$ as in \ac{ACQL}), and $\gamma_c \in (0, 1)$, is a contraction mapping.
    \begin{proof}
        \begin{align*}
            |g^c(Q^c)_{s,a} - g^c(\hat{Q}^c)_{s,a}| &= | \gamma_c \expectation_{s'} \left [ \min \{ c_{sa}, Q^c_{s',\pi(s')} \} \right ] - \gamma_c \expectation_{s'} \left [ \min \{ c_{sa}, \hat{Q}^c_{s', \pi(s')} \} \right ] | \\
            &= \gamma_c \expectation_{s'} \left [ |\min \{ c_{sa}, Q^c_{s',\pi(s')} \} - \min \{ c_{sa}, \hat{Q}^c_{s', \pi(s')} \} | \right ] \\
            &\leq \gamma_c \expectation_{s'} \left [ | Q^c_{s',\pi(s')} - \hat{Q}^c_{s', \pi(s')} | \right ] \quad {(\tiny |\min\{a,b\} - \min\{a,c\}| \leq |b - c|)} \\
            &\leq \gamma_c || Q^c - \hat{Q}^c ||_\infty
        \end{align*}
        Therefore, $||g^c(Q^c) - g^c(\hat{Q}^c)||_\infty \leq \gamma_c || Q^c - \hat{Q}^c ||_\infty$.
    \end{proof}
\end{lemma}

\begin{lemma}
    \label{lemma:qr-convergence-for-qc}
    As $n \rightarrow \infty$, $Q^c(n)$ converges to a fixed point $\lambda_1(Q^r, \gamma_c)$ for some fixed $Q^r$ and $\gamma_c$.
    \begin{proof}
        The convergence of $Q^c$ for a fixed $Q^r$ and $\gamma_c$ follows from Theorem 3 (convergence of distributed stochastic approximation algorithms for a contraction mapping) in \cite{tsitsiklis1994asynchronous}. Assumptions 1, 2, and 3 in \cite{tsitsiklis1994asynchronous} are satisfied due to our Assumptions~\ref{assumption:eventually-discarding-info}, \ref{assumption:robbins-monro-separated-step-sizes} and the definitions of $M^c_{s,a}$, $M^r_{s,a}$ in \eqref{eqn:mc-martingale-difference-sequence} and \eqref{eqn:mr-martingale-difference-sequence}. Furthermore, the contraction property of $g^c$ (Lemma~\ref{lemma:g-contraction-property}) is enough to satisfy Assumption 5 in \cite{tsitsiklis1994asynchronous}. Under these conditions, Theorem~3 in \cite{tsitsiklis1994asynchronous} holds true.
    \end{proof}
\end{lemma}
\begin{lemma}
    \label{lemma:qc-convergence}
    As $n \rightarrow \infty$, $Q^r(n)$ updated with \eqref{eqn:stochastic-qr-update} using a fixed $Q^c(n) = \lambda_1(Q^r(n), \gamma_c)$, such that there is a feasible action in every state, converges to the optimal value function $Q^{r^*}_{\gamma_c}$.
    \begin{proof}
        The mapping $h^r = h(\cdot, Q^c, \gamma_c) : \mathbb{R}^d \rightarrow \mathbb{R}^d$ for a fixed $Q^c = \lambda(Q^r, \gamma_c)$ is a typical Bellman operator for a fixed policy using only actions from a constant non-empty subset of $\mathcal{A}$ for each state $s$. As a result, Theorem 4 (convergence of standard Q-learning) in \cite{tsitsiklis1994asynchronous} applies.
    \end{proof}
\end{lemma}
\begin{lemma}
    \label{lemma:qr-qc-convergence-for-gamma-c}
    $Q^r(n)$ and $Q^c(n)$ asymptotically approach $Q^{r^*}_{\gamma_c}$ and $Q^{c*}_{\gamma_c} = \lambda_1(Q^{r^*}_{\gamma_c}, \gamma_c)$ as $n \rightarrow \infty$.
    \begin{proof}
        Lemmas \ref{lemma:qr-convergence-for-qc} and \ref{lemma:qc-convergence} serve to satisfy Assumptions 1 and 2 in Chapter 6 of \cite{borkar2008stochasticapproximation}. The boundedness of our rewards and constraint signals also result in a bounded $Q^r(n)$ and $Q^c(n)$, which satisfies Assumption 3 in Chapter 6 of \cite{borkar2008stochasticapproximation}. The proof follows from Theorem 2 (convergence of two-timescale coupled stochastic approximation algorithms) in the same chapter.
    \end{proof}
\end{lemma}

\subsection{Convergence of $(Q^r, Q^c)$ and $\gamma_c$}

We can apply a similar two-timescale argument now using $\gamma_c$ on the slower timescale and $(Q^r(n), Q^c(n))$ on the faster timescale. The condition that the faster timescale converges to a fixed point $\lambda_2(\gamma_c)$ for a static $\gamma_c$ is shown in Lemma~\ref{lemma:qr-qc-convergence-for-gamma-c}. For the condition that the slower timescale converges to a fixed point with $(Q^r(n), Q^c(n)) = \lambda_2(\gamma_c)$) is true trivially because $\gamma_c$ converges without depending on $(Q^r(n), Q^c(n))$ at all.
\begin{lemma}
    \label{lemma:gamma-c-convergence}
    $\gamma_c$ converges to $1$ as $n \rightarrow \infty$.
    \begin{proof}
        The update in \eqref{eqn:stochastic-gamma-c-update} is a discretization of the ODE $\dot{\gamma_c}(t) = 1 - \gamma_c(t)$. The solution to this ODE is $\gamma_c(t) = 1 - (1 - \gamma_c(0))e^{-t}$, which asymptotically approaches 1 as $t \rightarrow \infty$.
    \end{proof}
\end{lemma}

Finally, similar to Proposition~1 in \cite{fisac2019bridging}, we observe that $\lim_{\gamma_c \rightarrow 1} g^c_{s,a}(Q^c) = \min \{ c_{sa}, \expectation_{s'} Q^c_{s', \pi(s')} \} $ yields a fixed point at $Q^{c*}_{s,a} = \expectation_{\tau \sim \pi} \left [ \min_{t \in [0, \infty]} c_{{sa}_t} | s_0=s, a_0=a \right ]$ matching the undiscounted minimum-safety constraint (Equation \eqref{eqn:min-cost-cmdp} in our main paper). Using this fact and another application of Theorem 2 in \cite{borkar2008stochasticapproximation}, we can prove our Proposition~\ref{proposition:convergence-restated}.

\begin{proposition}
    \label{proposition:convergence-restated}
    Let $\mathcal{M^A}$ be an augmented \ac{CMDP} with $|\mathcal{S}^A| < \infty$, $|\mathcal{A}| < \infty$, and $\gamma \in [0, 1)$, and let $Q^c(n)$ and $Q^r(n)$ be models for the state-action safety and value functions indexed by $n$. Assume they are updated using Robbins-Monro step sizes $a(n)$ and $b(n)$, respectively, with $b(n) \in o(a(n))$ according to \eqref{eqn:qc-update} and \eqref{eqn:qr-update}. Assume that $\gamma_{c}(n)$ is also updated with step sizes $c(n)$ such that $\gamma_{c}(n) \rightarrow 1$ and $c(n) \in o(b(n))$. Then $Q^c(n)$ and $Q^r(n)$ converge to $Q^{c*}$ and $Q^{r*}$ almost surely as $n \rightarrow \infty$.
    \begin{proof}
        By Theorem~2 from Chapter~6 of \cite{borkar2008stochasticapproximation}, whose
        conditions are satisfied by Lemmas \ref{lemma:qr-qc-convergence-for-gamma-c}
        and \ref{lemma:gamma-c-convergence}, the coupled iterates $(\gamma_c(n), Q^r(n), Q^c(n))$ converge almost surely to a fixed point $(1, \lambda_2(1))$ with $\lambda_2(\gamma_c) = (Q^{r*}_{\gamma_c}, \lambda_1(Q^{r*}_{\gamma_c}, \gamma_c))$ as $n \to \infty$. As $\gamma_c \to 1$,  $Q^c(n) = \lambda_1(Q^{r*}_{\gamma_c}, \gamma_c)$ also converges to $ \expectation_{\tau \sim \pi} \left [ \min_{t \in [0, \infty]} c_{{sa}_t} | s_0=s, a_0=a \right ]$ with the policy $\pi$ determined by $\lim_{n \to \infty}\bar{Q}^{r}(n) = Q^{r*}$. Therefore, the algorithm attains the optimal state-action value function and the corresponding optimal (undiscounted) state-action safety function $Q^{c*}$.
    \end{proof}
\end{proposition}

\section{Additional \ac{ACQL} Details and Pseudocode}
\label{appendix:functions}

\paragraph{Automaton Analysis} 


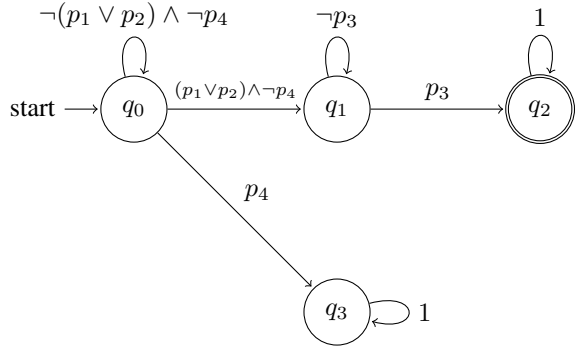
\begin{figure}
\centering
\begin{tikzpicture}[shorten >=1pt,node distance=2.7cm,on grid,auto] 
   \node[state,initial] (q_0)   {$q_0$}; 
   \node[state] (q_1) [right=of q_0] {$q_1$}; 
   \node[state,accepting](q_2) [right=of q_1] {$q_2$};
   \node[state](q_3) [below=of q_1] {$q_3$};
   \path[->] 
    (q_0) edge [sloped] node {$\scriptstyle (p_1 \vee p_2) \wedge \neg p_4$} (q_1)
          edge node {$p_4$} (q_3)
          edge [loop above] node {$\neg (p_1 \vee p_2) \wedge \neg p_4$} ()
    (q_1) edge [sloped] node {$p_3$} (q_2)
          edge [loop above] node {$\neg p_3$} ()
    (q_2) edge [loop above] node {$1$} ()
    (q_3) edge [loop right] node {$1$} ();
\end{tikzpicture}
\caption{Automaton for the task ``Reach goal $g_1$ or $g_2$ while never entering an unsafe-region $u_1$. Then reach $g_3$.'', where achieving the goal $g_i$ corresponds to the atomic proposition $p_i$ and entering $u_1$ corresponds to $p_4$. The full \ac{LTL} expression is $\neg p_4 \; \mathcal{U} \; ((p_1 \vee p_2) \wedge \circ \lozenge p_3)$. The proposition $p_4$ is only relevant to the task's safety constraint, and the propositions $p_1$, $p_2$, and $p_3$ are only relevant to the task's liveness constraints.}
\label{fig:automaton-example}
\end{figure}

To better illustrate the initial automaton analysis in \ac{ACQL} (Line 2), consider the automaton in Figure~\ref{fig:automaton-example}. There is only one non-accepting sink-components of this automaton, and it is the component consisting of the single node $q_3$. For $q_0$, there is only one transition into this component via the edge labeled by $p_4$, so the safety for $q_0$ condition is $S(q_0) = \neg p_4$. For $q_1$ and $q_2$, there are no transitions to a non-accepting sink-component and so their safety conditions are $S(q_1) = S(q_2) = 1$, meaning that the safety condition is trivially satisfied at all times in those states. For completeness, we also consider the unsafe states themselves as having safety conditions equal to the conjunction of their negated incoming transitions. Therefore, $q_3$ also has the safety condition $S(q_3) = \neg p_4$. 

Now we can obtain the liveness constraints, which are summarized in the liveness condition mapping $O : \mathcal{Q} \rightarrow \Phi$ (See Section \ref{section:obtaining-g} in our main paper). For $q_0$, the only remaining outgoing edge is the one labeled $\neg (p_1 \vee p_2) \wedge \neg p_4$. Since $S(q_0) = \neg p_4$, we can eliminate it from this transition predicate to obtain $O(q_0) = (p_1 \vee p_2)$. For $q_1$, one can simply obtain $O(q_1) = p_3$ from its only outgoing edge. For completeness, we also set $O(q_2) = p_3$ using the incoming edges for $q_2$ since it is an accepting state. This is purely to inform the agent of the goal associated with an accepting state once it has reached it. From the above values for $O$, the mapping $G$ can be defined by $G(q_0) = \{ g_1, g_2 \}$, $G(q_1) = \{ g_3 \}$ and $G(q_2) = \{ g_3 \}$.

\paragraph{Subroutines}

Algorithms~\ref{alg:get-trajectory}, \ref{alg:relabel}, and \ref{alg:safety_gamma_scheduler} present the pseudo-code for collecting trajectories in $\mathcal{M}^A$, relabeling them with achieved goals, and computing the safety discount factor $\gamma_c$.

The GetTrajectory function, in Algorithm~\ref{alg:get-trajectory}, begins by sampling an initial state from the distribution $d^A_0$. We also randomly sample positions from the environment to associate with each subgoal proposition $p \in AP_\text{subgoal}$. This task randomization promotes collecting trajectories that explore a greater portion of the state space through a variety of subgoal sequences, and we found that this was necessary to stabilize training while using \ac{HER}. We speculate that the greater variety of state and subgoal combinations is needed to train a robust subgoal-reaching policy. The agent proceeds to interact with the environment for a total of $T$ steps. In a loop indexed by $t$, actions are selected from an epsilon-greedy version of the input policy $\pi$ (Line 5). The reward, constraint feedback, and next state are observed (Lines 6-8) and stored in the trajectory $\tau$ (Line 9). After $T$ steps have executed, the trajectory is returned.


The Relabel function, in Algorithm~\ref{alg:relabel}, takes a batch of trajectories $\mathcal{B}_\tau$. For each trajectory in the batch, the function determines the final achieved state $g'$ and overwrites a single subgoal for every step in the entire trajectory with $g'$. It also overwrites the reward for each step with 1 for steps that were sufficiently close to $g'$ and 0 for steps that were not. We found that this relatively simple strategy, despite the fact that trajectories were collected with multiple-subgoals in-mind, was sufficient to train reliable goal-reaching policies.

The SafetyGammaScheduler function, in Algorithm~\ref{alg:safety_gamma_scheduler}, generates values for $\gamma_c$, starting from an initial value and gradually increasing toward $1.0$ with exponential decay. To ensure accurate learning of state-action safety function models, it was necessary to cap $\gamma_c$ at a value slightly below $1.0$.


\begin{algorithm}[t]
\small
\caption{GetTrajectory} \label{alg:get-trajectory}
\begin{algorithmic}[1]
    \Function{GetTrajectory}{$\mathcal{M}^A$, $\pi$}
        \State $s^A_0 \sim d_0^A$, $g_i \sim P(\mathcal{G})$ for $p_i \in AP_{\text{subgoals}}$
        \State $t \leftarrow 0$, $\tau \leftarrow ()$
        \While{$t < T$}
            \State $a_t \sim \pi_{\epsilon-\text{greedy}}(s^A_t)$
            \State $r_t \leftarrow r^A(s^A_t, a_t)$
            \State $c_t \leftarrow c^A(s^A_t, a_t)$
            \State $s^A_{t+1} \sim \mathcal{T}^A(s^A_t, a_t)$
            \State $\tau \leftarrow \tau \cup (s^A_t, r_t, c_t, a_t)$
            \State $t \leftarrow t + 1$
        \EndWhile
        \State \Return $\tau$
    \EndFunction
\end{algorithmic}
\end{algorithm}

\begin{algorithm}[t]
\small
\caption{Relabel} \label{alg:relabel}
\begin{algorithmic}[1]
    \Function{Relabel}{$\mathcal{B}_r$}
        \For{$\tau \in \mathcal{B}_r$}
            \State $g' \leftarrow $ the final goal state achieved in $\tau$.
            \For{$s^A_t, r_t \in \tau$}
                \State Replace $g_1$ in $s^A_t$ with $g'$
                \State $r_t \leftarrow 1$ if $g'$ achieved in $s^A_t$ and $0$ otherwise.
            \EndFor
        \EndFor
        \State \Return $\mathcal{B}_r$
    \EndFunction
\end{algorithmic}
\end{algorithm}

\begin{algorithm}[t]
\small
\caption{SafetyGammaScheduler} \label{alg:safety_gamma_scheduler}
\begin{algorithmic}[1]
    \Function{SafetyGammaScheduler}{$j$}
        \State $\text{x} \leftarrow j \div \text{update\_period}$
        \State $\text{y} \leftarrow 1.0 - (1.0 - \text{init\_value}) \cdot \text{decay\_rate}^{\text{x}}$
        \State \Return $\begin{cases}
            \text{max\_value} & \text{if} \; \text{y} \geq \text{max\_value} \\
            \text{y} & \text{o.w.}
        \end{cases}$
    \EndFunction
\end{algorithmic}
\end{algorithm}

\section{Model Architectures, Environment Implementation Details, and Hyper-Parameters}
\label{appendix:implementation-details}
\label{appendix:environment-details}

\paragraph{Model Architecture and Policy} Our method trains two models: the state-action value function, $Q^r_\psi$, and the state-action safety function, $Q^c_\theta$. Both models employ twin neural networks and use the minimum of their predicted values to mitigate overestimation in value and safety estimation. For the ablation study using a state-action sum-of-costs function, we likewise use twin networks, but instead take the maximum of the two predictions to maintain a conservative (i.e., pessimistic) estimate.

To simplify handling multiple goals, $Q^r_\psi$ is defined using a goal-conditioned state-action value function model, $Q^{GC}_\psi : \mathcal{S} \times \mathcal{G} \times \mathcal{Q} \times \mathcal{A} \rightarrow \mathbb{R}$, parameterized by $\psi$. At runtime, $Q^{GC}_\psi$ is called for each subgoal $g \in g^+$. This approach exploits the fact that goal-conditioned value functions form a Boolean algebra under the $\min$ and $\max$ operators \cite{tasse2020booleantaskalgebrareinforcement, tasse2022skill}. For example, consider two subgoal propositions $p_1, p_2 \in AP_\text{subgoal}$. In the simplest case, when $O(q) = p_1$, we compute $Q^r_\psi(\langle s, g_1, q \rangle, a)$ as $Q^{GC}_\psi(s, g_1, q, a)$. When $O(q) = p_1 \wedge p_2$ (i.e., both subgoals must be achieved to progress), we compute the value as the minimum of $Q^{GC}_\psi(s, g_1, q, a)$ and $Q^{GC}_\psi(s, g_2, q, a)$. Conversely, when $O(q) = p_1 \vee p_2$ (i.e., achieving either subgoal suffices), the value is the maximum of $Q^{GC}_\psi(s, g_1, q, a)$ and $Q^{GC}_\psi(s, g_2, q, a)$. This pattern generalizes to arbitrarily complex Boolean formulae, allowing us to efficiently approximate $Q^r_\psi$ using a single goal-conditioned network.

We also observed that, because liveness constraints are separated into the goal input, conditioning behavior on the automaton state is only necessary when safety constraints differ between automaton states. As a result, we do not require a distinct “mode” for every automaton state $q \in \mathcal{Q}$, but only for each unique safety condition in the mapping $S$. To encode this, we train $Q^r_\psi$ using a multi-headed neural network, where each output head corresponds to a distinct safety condition. For example, in the automaton shown in Figure~\ref{fig:automaton-example}, the mapping $S$ assigns states to one of two safety conditions: $\neg p_4$ or $1$. Accordingly, $Q^{GC}_\psi$ has two output heads, selected based on the currently active safety condition. In all \ac{ACQL} experiments, $Q^{GC}_\psi$ shared a hidden layer of 256 neurons across all heads; each head then had an additional hidden layer of 256 neurons. All layers used ReLU activations, and the final output layer had $|\mathcal{A}|$ neurons with no activation function.

The model $Q^c_\theta$ follows the same architecture as $Q^r_\psi$, with a goal-conditioned, multi-headed neural network, but differs in two key respects. First, for disjunctive goal conditions ($p_1 \vee p_2$), the output is defined as the minimum of $Q^{GC}_\theta(s, g_1, q, a)$ and $Q^{GC}_\theta(s, g_2, q, a)$, to ensure conservative safety by taking the worst-case estimate across disjunctive paths. Second, $Q^{GC}_\theta$ uses a different network architecture: it has two shared hidden layers of 64 neurons each, and each output head includes two additional hidden layers with 64 and 32 neurons, respectively. All layers use ReLU activations. The final output layer consists of $|\mathcal{A}|$ neurons with a $\tanh$ activation function.

The policy was implemented in terms of these two value functions according to the constrained maximization $\pi(s) = \argmax_{a : Q^c_\theta(s, a) > \mathcal{L}}Q^r_\psi(s, a)$. In the case that no action was deemed feasible by $Q^c_\theta$, the safest action $\max_a Q^c(s, a)$ was chosen. The exploration policy $\pi_{\epsilon-\text{greedy}}$ would behave as above with probability $1 - \epsilon$, and with probability $\epsilon$ select a random action without considering $Q^r$ or $Q^c$.

\paragraph{Environment Implementation} For our simulated experiment environments, we used the Brax physics simulator \cite{brax2021github} and assets provided in JaxGCRL \cite{bortkiewicz2024accelerating} for the PointMass, Quadcopter, and Ant environments. Acting in these environments was facilitated by a set of discrete actions corresponding to movement in the cardinal directions. Specifically, the actions in the PointMass and Quadcopter environments output a constant low-level action to accelerate in one of the four or six available directions for 5 consecutive steps. The actions for our UR5e environment, which we used to train our real-world-deployed policies, similarly moved the robot's end effector in the 6 cardinal directions for a single time step. The actions in the AntMaze environment were policies trained separately using \ac{PPO} \cite{schulman2017proximalpolicyoptimizationalgorithms} for 50000000 environment interactions with the objective of maximizing velocity in each of the four cardinal directions and would run for 4 consecutive steps when executed. All further details regarding environment geometry and task definitions for our simulated and real-world experiments are included in our \href{https://github.com/Tass0sm/acql}{Code Repository}\footnote{\label{repo}https://github.com/Tass0sm/acql}.


\paragraph{Hyper-Parameters} Table~\ref{tab:hyperparameters} reports the hyperparameter values most commonly used in our experiments, including hyperparameters for the safety gamma ($\gamma_c$) scheduler described in Appendix~\ref{appendix:functions}. For a complete account of hyperparameters, as well as \ac{ACQL}, baseline, and environment implementation details, refer to our \href{https://github.com/Tass0sm/min-max-rl}{Code Repository}\footnotemark{\ref{repo}}.

\begin{table}[H]
\centering
\caption{Hyperparameter values used for experiments in Tables~\ref{tab:task-experiments} and \ref{tab:ablations} in our main paper}
\label{tab:hyperparameters}
\small
\begin{tabular}{c c c c}\toprule
\multicolumn{2}{l}{Hyperparameter Name} & \multicolumn{2}{c}{Value}\\
\cmidrule(lr){1-2} \cmidrule(lr){3-4}

\multicolumn{2}{l}{Episode length $(T)$} & \multicolumn{2}{c}{$1000$} \\
\multicolumn{2}{l}{Discount factor $(\gamma)$} & \multicolumn{2}{c}{$0.99$} \\
\multicolumn{2}{l}{Learning rate $(\alpha)$} & \multicolumn{2}{c}{$1\cdot10^{-4}$} \\
\multicolumn{2}{l}{$\epsilon$-greedy factor $(\epsilon)$} & \multicolumn{2}{c}{$0.1$} \\
\multicolumn{2}{l}{Safety limit $(\mathcal{L})$} & \multicolumn{2}{c}{$0.0$} \\
\multicolumn{2}{l}{Safety Gamma Init Value} & \multicolumn{2}{c}{$0.80$} \\
\multicolumn{2}{l}{Safety Gamma Update Period} & \multicolumn{2}{c}{$250,000$} \\
\multicolumn{2}{l}{Safety Gamma Decay Rate} & \multicolumn{2}{c}{$0.15$} \\
\multicolumn{2}{l}{Safety Gamma Max. Value} & \multicolumn{2}{c}{$0.98$} \\

\multicolumn{2}{l}{Target parameter interpolation factor $(\lambda)$} & \multicolumn{2}{c}{$0.005$}\\
\bottomrule
\end{tabular}
\end{table}



\paragraph{Compute Resource Requirements}
\label{appendix:compute}

All experiments were conducted on a single NVIDIA RTX 3090 GPU (24 GB VRAM), using a local workstation equipped with an 12th Gen Intel i7-12700F CPU, 32 GB RAM. No cloud services or compute clusters were used. Each individual experimental run required approximately 30 minutes of compute time on the GPU. The full experiment grid consists of 225 runs for the comparative analysis and 90 runs for the ablations, amounting to approximately 315 GPU-hours. Minor additional compute was used for initial hyperparameter tuning and development.

\section{Expanded Experimental Results}

\begin{figure*}[]
    \centering
    \includegraphics[width=0.9\linewidth]{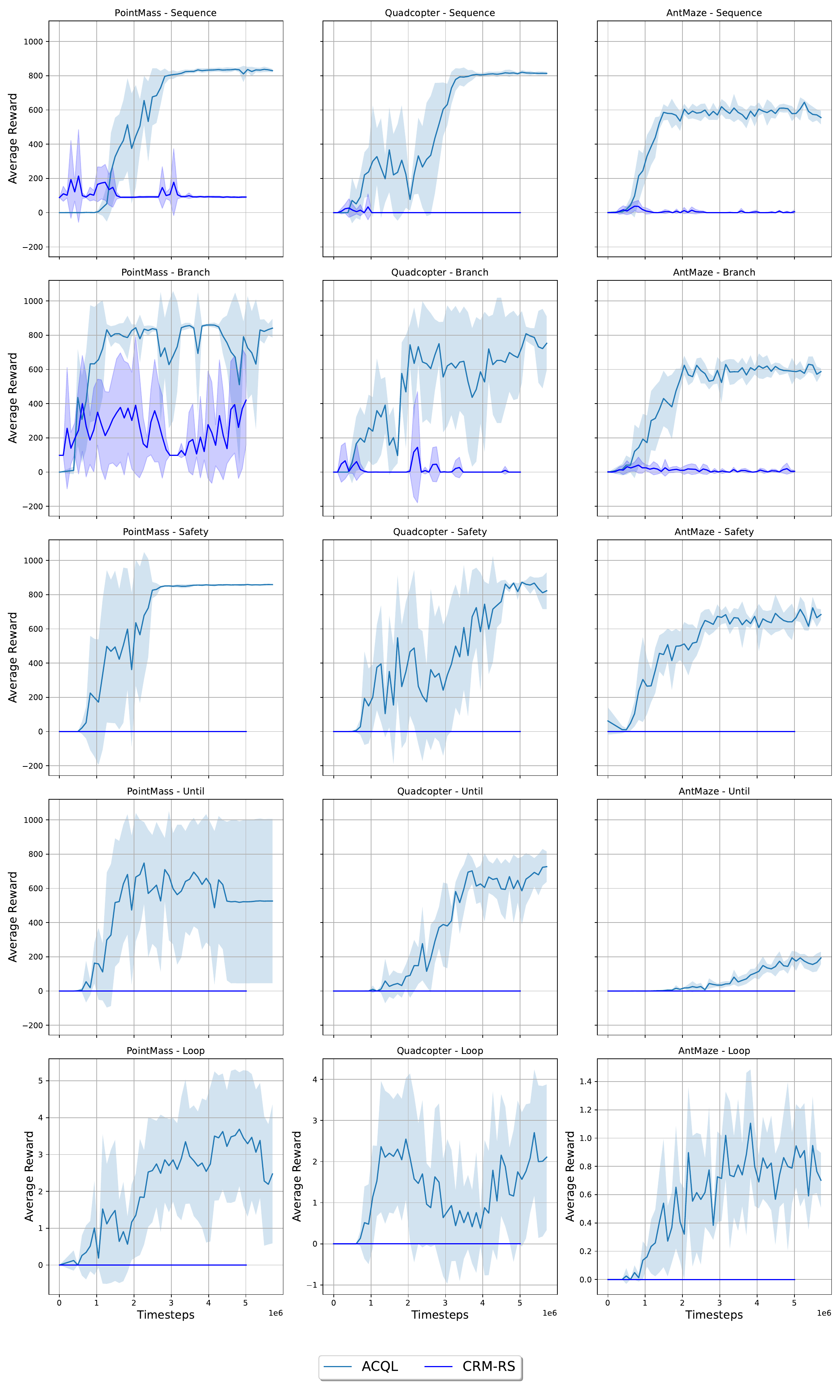}
    \caption{Average and one standard deviation of episode reward throughout training for the five runs per method that are summarized in Table~\ref{tab:task-experiments} in our main paper.}
    \label{fig:training-reward-plot}
\end{figure*}

\begin{figure*}[]
    \centering
    \includegraphics[width=0.9\linewidth]{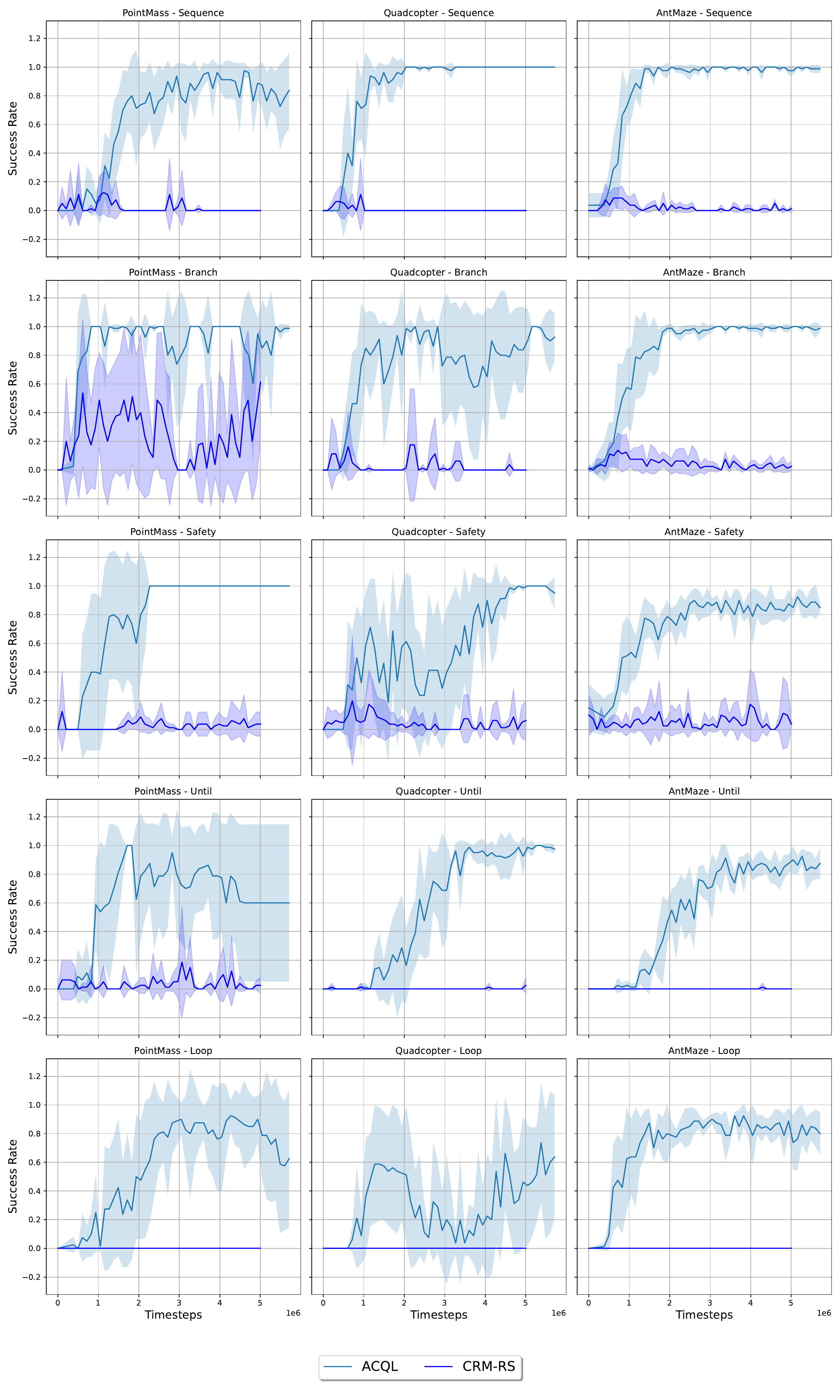}
    \caption{Average and one standard deviation of episode success rate throughout training for the five runs per method that are summarized in Table~\ref{tab:task-experiments} in our main paper.}
    \label{fig:training-success-rate-plot}
\end{figure*}

\begin{figure*}[]
    \centering
    \includegraphics[width=\linewidth]{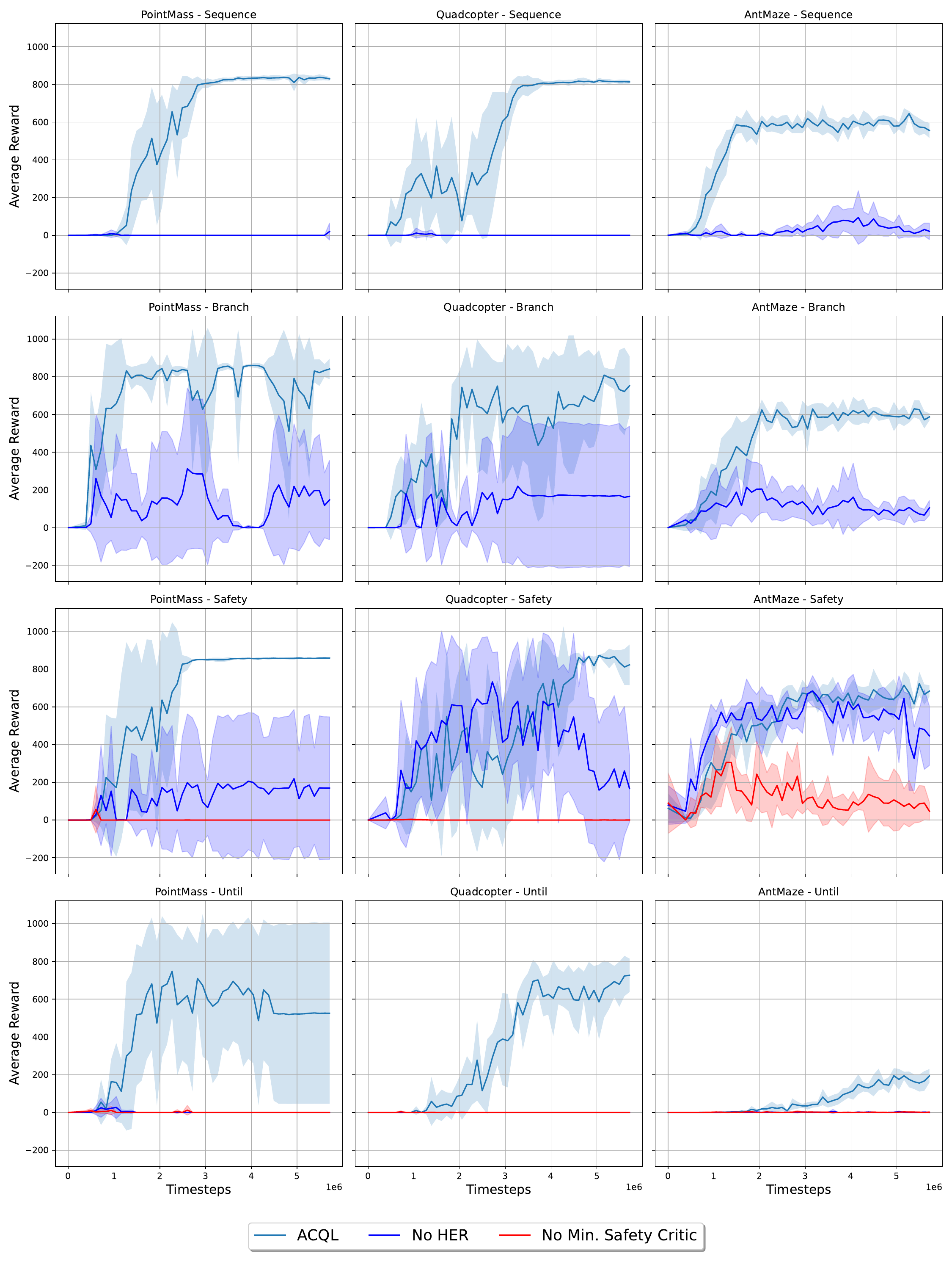}
    \caption{Average and one standard deviation of episode reward throughout training for the five runs per ablation group that are summarized in Table~\ref{tab:ablations} in our main paper.}
    \label{fig:ablation-training-reward-plot}
\end{figure*}

\begin{figure*}[]
    \centering
    \includegraphics[width=\linewidth]{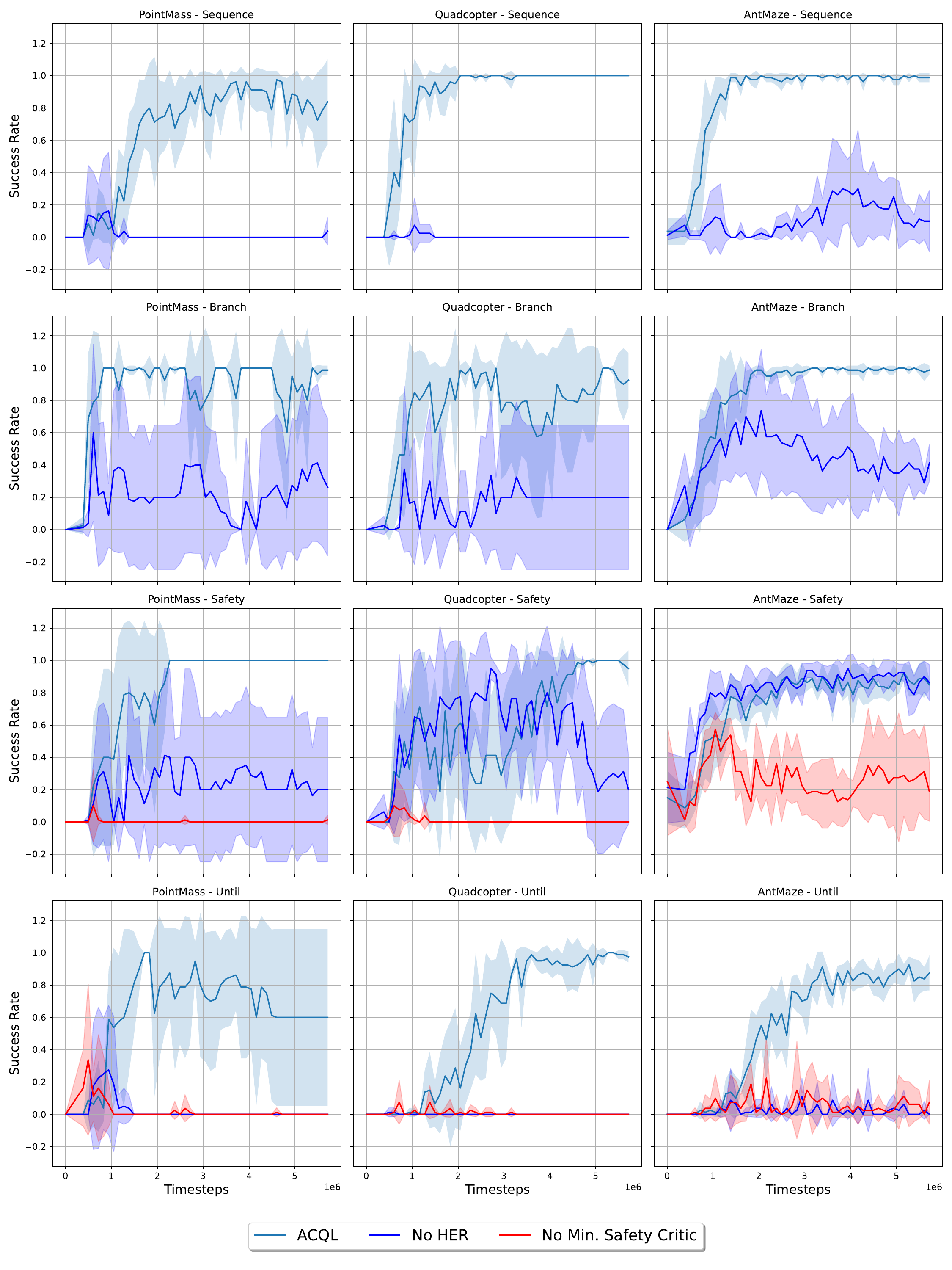}
    \caption{Average and one standard deviation of episode success rate throughout training for the five runs per ablation group that are summarized in Table~\ref{tab:ablations} in our main paper.}
    \label{fig:ablation-training-success-rate-plot}
\end{figure*}

Figures~\ref{fig:training-reward-plot} and \ref{fig:training-success-rate-plot} show the average reward and success rate throughout training for the baseline comparison experiments summarized in Table~\ref{tab:task-experiments}. The \ac{LOF} baseline cannot be depicted on these plots as it does not learn a single policy in the same \ac{MDP} as the other methods, and instead learns a policy that chooses subgoal-specific options for an abstracted state space $\mathcal{Q} \times \mathcal{S}_g$ constructed from the automaton states $\mathcal{Q}$ and the finite set of states $\mathcal{S}_g \subset \mathcal{S}$ corresponding to task subgoals. Figures~\ref{fig:ablation-training-reward-plot} and \ref{fig:ablation-training-success-rate-plot} show the average reward and success rate throughout training for the experiments summarized in Table~\ref{tab:ablations} in our main paper. The differences in amount of training steps depicted by the figures is due to the different design and training pipelines that the two algorithms observe. \ac{ACQL} collects complete trajectories to store in the Replay Buffer and \ac{CRM-RS} just collects individual transitions. We want to highlight that our algorithm converges earlier during the training and this difference does not play a significant role in the performance gap reported in Table~1.

\clearpage


\end{document}